\documentclass[runningheads]{llncs}

\usepackage{eccv}

\usepackage{setspace}
\usepackage{graphicx}
\usepackage{booktabs}
\usepackage{amssymb}
\usepackage{makecell}
\usepackage{verbatim}
\usepackage[ruled, lined, boxed]{algorithm2e}
\usepackage{multirow}
\usepackage{enumitem}
\usepackage{microtype}
\usepackage{bbding}

\usepackage{threeparttable} 

\usepackage{algorithmic}

\usepackage[utf8]{inputenc} % allow utf-8 input
\usepackage[T1]{fontenc}    % use 8-bit T1 fonts
\usepackage{url}            % simple URL typesetting
\usepackage{booktabs}       % professional-quality tables
\usepackage{amsfonts}       % blackboard math symbols
\usepackage{nicefrac}       % compact symbols for 1/2, etc.
\usepackage{microtype}      % microtypography
\usepackage{xcolor}         % colors

\usepackage{eccvabbrv}

\usepackage{graphicx}
\usepackage{booktabs}

\usepackage[accsupp]{axessibility}  % Improves PDF readability for those with disabilities.

\usepackage[pagebackref,breaklinks,colorlinks,citecolor=eccvblue]{hyperref}
\usepackage{orcidlink}

\begin{document}

% \title{UniHDA: Towards Universal Hybrid Domain Adaptation of Image Generators} 
\title{UniHDA: A Unified and Versatile Framework for Multi-Modal Hybrid Domain Adaptation}

\titlerunning{UniHDA}

% % TODO FINAL: Replace with your author list. 
% % Include the authors' OCRID for the camera-ready version, if at all possible.
% Second Author\inst{2,3}\orcidlink{1111-2222-3333-4444} \and
% Third Author\inst{3}\orcidlink{2222--3333-4444-5555}}

% % TODO FINAL: Replace with an abbreviated list of authors.
\authorrunning{Hengjia Li et al.}

% % TODO FINAL: Replace with your institution list.
% Springer Heidelberg, Tiergartenstr.~17, 69121 Heidelberg, Germany

\author{Hengjia Li$^{1}$ \and Yang Liu$^{1}$ \and Yuqi Lin$^{1}$\and Zhanwei Zhang$^{1}$\and Yibo Zhao$^{1}$\and Weihang Pan$^{1}$\and Tu Zheng$^{2}$\and Zheng Yang$^{2}$\and Chunjiang Yu$^{3}$\and Boxi Wu$^{1}$\and Deng Cai$^{1}$ 
}
\institute{
\textsuperscript{1} {Zhejiang University} \;
\textsuperscript{2} {Fabu Inc.} \;
\textsuperscript{3} {Ningbo Port} \;
}

\maketitle

\begin{center}
    \centering
    \includegraphics[width=0.99\linewidth]{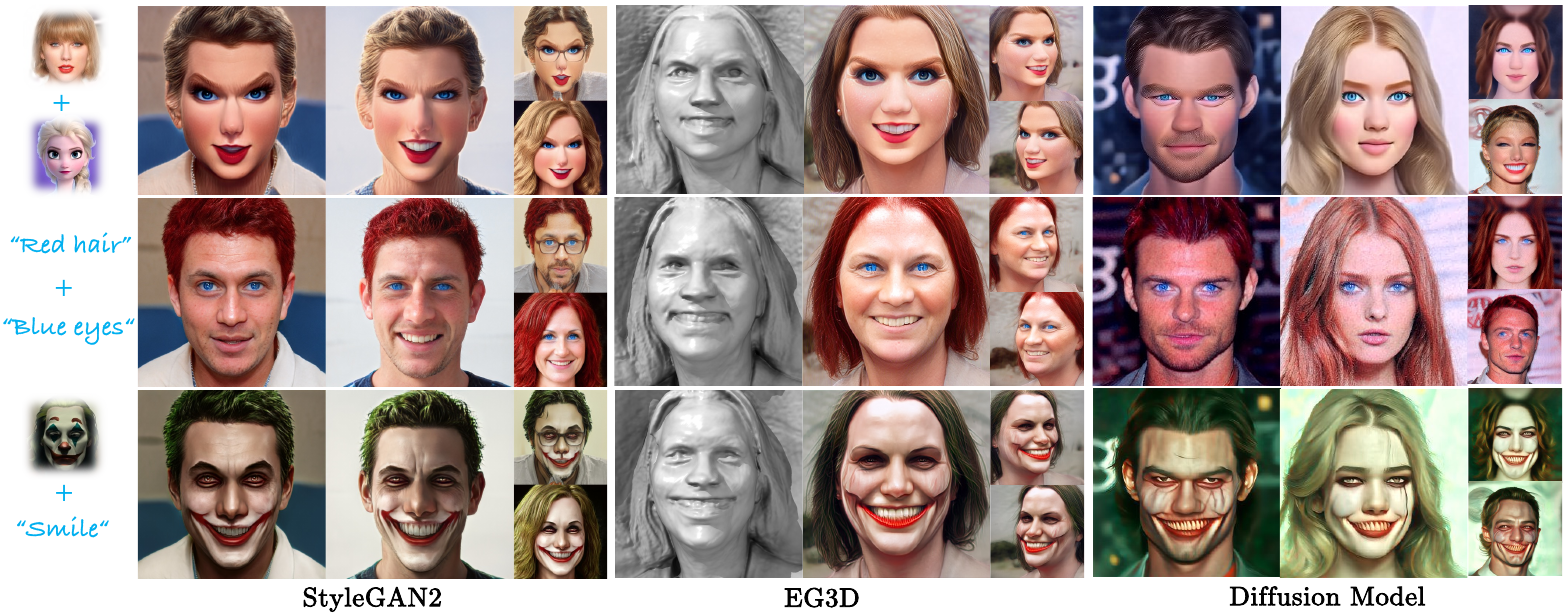} 
    \vspace{-0.2cm}
    \captionof{figure}{Given a pre-trained source generator and multiple target domains, UniHDA adapts the generator to a hybrid target domain that blends all characteristics at once and maintains robust cross-domain consistency. UniHDA supports both image and text modalities and is versatile to multiple generators.
    }
    \vspace{-0.3cm}
    \label{fig:task}
\end{center}

\begin{abstract}
Recently, generative domain adaptation has achieved remarkable progress, enabling us to adapt a pre-trained generator to a new target domain. However, existing methods simply adapt the generator to a single target domain and are limited to a single modality, either text-driven or image-driven. Moreover, they cannot maintain well consistency with the source domain, which impedes the inheritance of the diversity.
% Moreover, they are prone to overfitting domain-specific attributes, which inevitably compromises cross-domain consistency.
In this paper, we propose UniHDA, a \textbf{unified} and \textbf{versatile} framework for generative hybrid domain adaptation with multi-modal references from multiple domains. We use CLIP encoder to project multi-modal references into a unified embedding space and then linearly interpolate the direction vectors from multiple target domains to achieve hybrid domain adaptation.
To ensure \textbf{consistency} with the source domain, we propose a novel cross-domain spatial structure (CSS) loss that maintains detailed spatial structure information between source and target generator. Experiments show that the adapted generator can synthesise realistic images with various attribute compositions. Additionally, our framework is generator-agnostic and versatile to multiple generators, \eg, StyleGAN, EG3D, and Diffusion Models.
\keywords{Generative Domain Adaptation \and Multi-Modal Adaptation \and Hybrid Domain Adaptation \and Generative Models}
\end{abstract}    
\section{Introduction}
\label{sec:intro}

\begin{figure*}[t]
\centering
\includegraphics[width=0.95\linewidth]{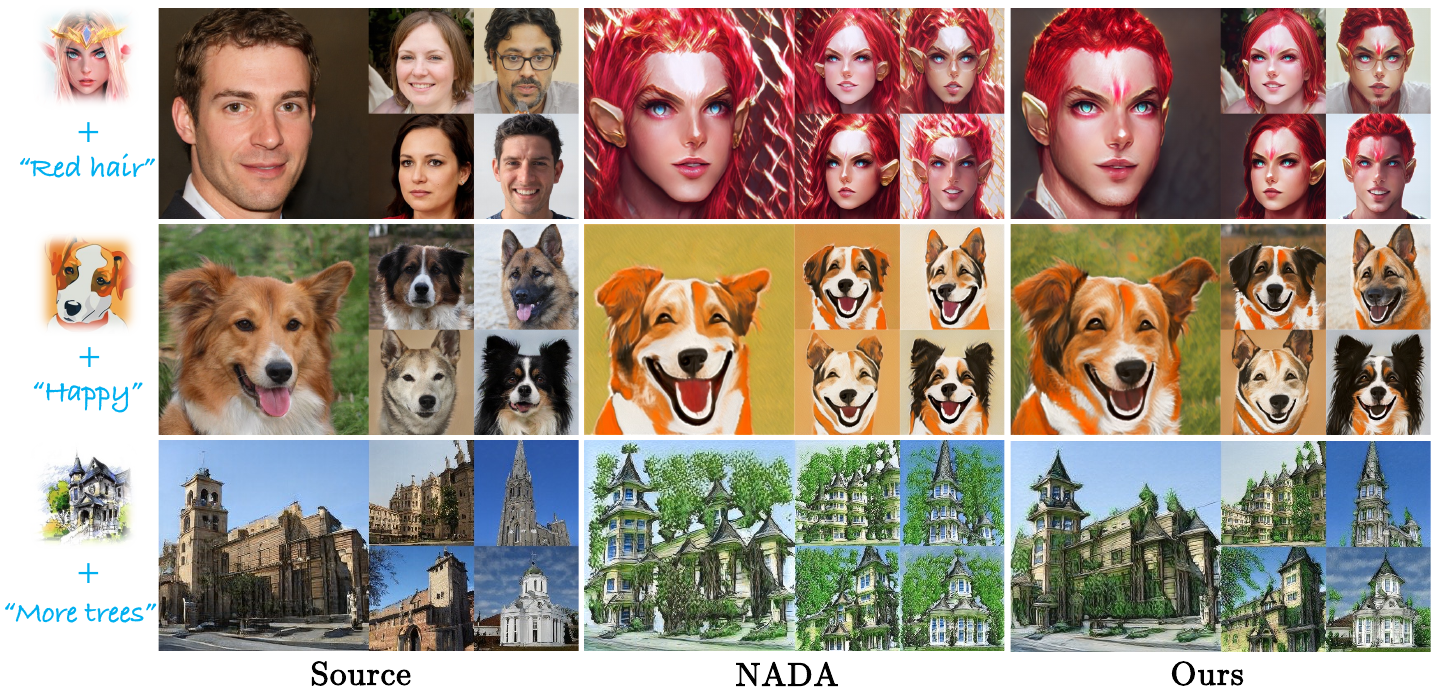}
\vspace{-0.2cm}
\caption{Existing methods like NADA~\cite{gal2021stylegan} fail to maintain consistency with the source domain for hybrid domain adaptation, resulting in overfitting to the limited references and impeding the inheritance of the diversity in the source domain.
}
\label{fig:overfit}
\vspace{-0.5cm}
\end{figure*}

\begin{figure*}[t]
\centering
\includegraphics[width=0.95\linewidth]{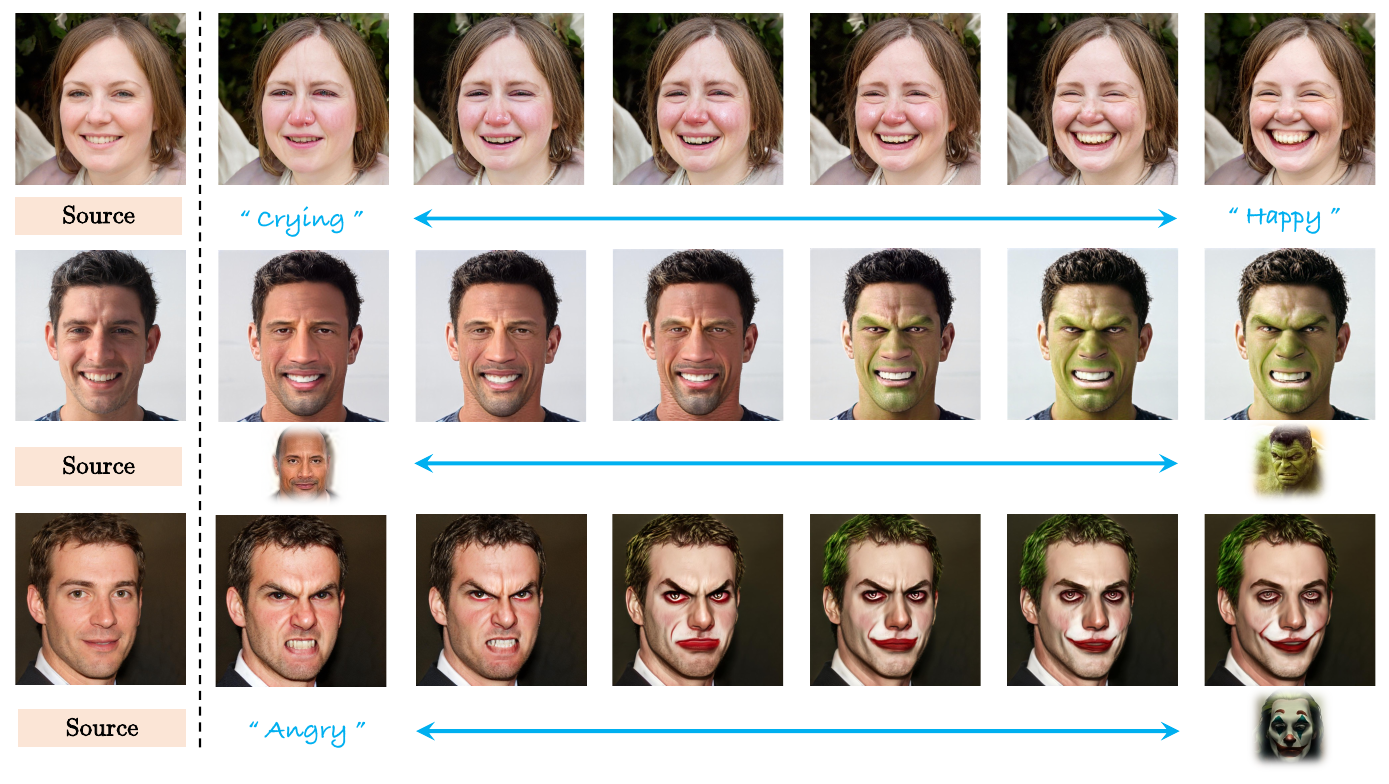}
\vspace{-0.2cm}
\caption{Linear interpolation between multi-modal direction vectors.
We represent the domain shift by the direction vector from source embedding to the target (\eg, \textit{Crying} or \textit{Happy}). Linear interpolation of them during training will result in a smooth traversal. The coefficients for the right domain are respectively 0, 0.2, 0.4, 0.6, 0.8, and 1, while for the left domain, they are set inversely.
% Linear interpolation of direction vectors. We can interpolate the direction vectors corresponding to different domains in CLIP's embedding space, resulting in a smooth adaptation as their coefficients change.
% The interpolation of the direction vectors in CLIP's embedding space will result in a smooth adaptation between different domains.(\eg, between source domain and \textit{crying} domain)
}
\label{fig:linear}
\vspace{-0.4cm}
% as the corresponding domain coefficients change  (\cref{eq:linear})
\end{figure*}

\quad Benefiting from tremendous success of modern image generator~\cite{karras2019style,brock2018large,vahdat2021score,rombach2022high}, generative domain adaptation has achieved remarkable progress during the past few years. Typically, it aims to adapt a pre-trained generator to a new target domain while preserving the variation in the source domain, \eg, from the \textit{human} domain to the \textit{baby} domain. Depending on the modality of references, generative domain adaptation can be categorized into two schools: text-driven~\cite{gal2021stylegan, YotamNitzan2023DomainEO, Liu_2023_ICCV, song2022diffusion, lei2023diffusiongan3d, kim2023datid, kim2023podia, alanov2022hyperdomainnet, zhu2022one, lyu2023deltaedit} and image-driven~\cite{mo2020freezeD, li2020fig_EWC, ojha2021fig_cdc, zhao2022closer, xiao2022few, mondalfew, wu2023domain, zhao2022few, zhao2023exploring, zhang2022towards, zhu2021mind, Alanov_2023_ICCV, duan2023weditgan, ye2023ip,kim2022diffface}. 

% In the pursuit of the versatility for target domains, HDA proposes image-driven hybrid domain adaptation to generate images with integrated attributes from individual target domains. 
Despite their promising results, there are still some limitations. A major limitation is that they only support adaptation from a source domain to individual target domain. These methods fail to directly adapt a \textit{human} generator to more practical real-world scenarios like \textit{person with red hair and blue eyes} or a combination of \textit{Taylor} and \textit{Elsa} (\cref{fig:task}). For more general purposes, previous methods also fail with multi-modal references, \eg, \textit{smiling Joker} given the \textit{non-smiling Joker} image and the \textit{smile} text.

Another limitation of existing approaches is that, they are prone to overfit domain-specific attributes, especially in hybrid domain adaptation (\cref{fig:overfit}). The main reason is that they do not maintain well consistency with the source domain, \eg, posture and identity. This results in overfitting to the limited references and losing the diversity in the source domain.

% With these techniques, we can acquire an adapted model to generate images in the target domain. However, various downstream application scenarios complicates the adaptation, including the situations where one model to support the composition of multiple target domains.
% % 引用stylemix， composer，magicmix
% Unfortunately, existing works are not sufficient to meet this demand. 
% emperically
% Under existing methods, one attribute often overwhelms the others in the adaptation process, which makes it difficult to retain the attributes of both domains simultaneously.

In this paper, we propose UniHDA, a \textbf{Uni}fied and versatile framework for generative \textbf{H}ybrid \textbf{D}omain \textbf{A}daptation with multi-modal references from multiple domains. UniHDA facilitates the references of individual image and text prompt simultaneously and blends the attributes from target domains to create a hybrid domain. To enable multiple modalities, we leverage pre-trained CLIP~\cite{radford2021learning} to project multi-modal references into a unified embedding space and represent the domain shift by the direction vector from the source embedding to the target embeddings. 

To achieve hybrid domain adaptation, we draw inspiration from the compositional capabilities in the latent space of StyleGAN~\cite{harkonen2020ganspace, shen2021closed, xu2022mind}.
%and explore the direction vector in CLIP's embedding space. 
We demonstrate that a semantically meaningful linear interpolation between direction vectors in CLIP's embedding space can uncover favorable compositional capabilities (\cref{fig:linear}). In light of this intriguing finding, we linearly interpolate direction vectors of multiple target domains to obtain the direction vector corresponding to the hybrid domain that semantically integrates attributes from all target domains.
% We find that the distinct direction vectors corresponding to different target domains only modify one aspect of synthesized images without altering other attributes, exhibiting favorable disentanglement. In light of this intriguing finding, we modulate the direction vectors from multiple target domains to seek out the direction vector corresponding to the hybrid domain.
% As an example shown in fig2, the modulated direction vector guides the model to generate images with integrated attributes from multiple target domains.

Furthermore, we introduce a novel cross-domain spatial structure loss (CSS) to preserve the consistency between source and target generator by maintaining detailed spatial structure information.
% Instead of CLIP, we leverage pre-trained DINO-ViT~\cite{dosovitskiy2020image, oquab2023dinov2} as the encoder, which is trained to focus on the distinction between subjects of the same class~\cite{ruiz2023dreambooth}.
Concretely, we leverage pre-trained Dino-ViT~\cite{dosovitskiy2020image, oquab2023dinov2} to encode generated images into patch tokens with fine-grained spatial information. For cross-domain consistency, we maintain the correspondence between source and target tokens with contrastive learning~\cite{oord2018representation}.
% Drawing inspiration from previous text-driven work~\cite{gal2021stylegan}, we utilize CLIP~\cite{radford2021learning} to represent the domain shifts by the direction of CLIP's embedding for images and texts. 
% % 最近的工作发现在文生图的diffusion模型里具有节藕能力 to some extent, 使得模型不需要finetune就能
% % 如果没有实验就用文字描述distanglement 如uncover
% Drawing inspiration from the disentanglement ability in the condition of text-to-image models~\cite{wu2023uncovering}, we find that disentanglement emerges through the embedding semantic space of CLIP~\cite{radford2021learning}, which facilitates us to intuitively control different attributes from multiple target domains in a linearly combinable manner. In light of this intriguing finding, we represent the domain shifts by the difference of CLIP's textual or visual embeddings. Then we modulate the shifts with pre-defined weights corresponding to multiple domains and sum them as the direction to guide the adaptation. %similar to NADA~\cite{gal2021stylegan}.
% Taking advantage of the disentanglement ability, the adapted model has the ability to generate images with attributes from multiple target domains while preserve their original characters from the source domain.  

% Furthermore, equipped with the framework, we can easily extend the paradigm to multi-modal references. As shown in fig1,....

% In addition, we show that this paradigm based on CLIP' embedding guidance is agnostic to pre-trained generative models, and present competitive results for image generation and editing of real and synthetic images.
We conduct experiments for a wide range of source and target domains to validate the effectiveness of our method. Results demonstrate that the adapted generator can synthesise realistic images with various attribute compositions.
In addition, we show that UniHDA is agnostic to the type of generators, \eg, StyleGAN~\cite{karras2019style, karras2020analyzing, karras2021alias}, EG3D~\cite{chan2022efficient}, and Diffusion models~\cite{Ho_Jain_Abbeel_2020, kim2022diffusionclip}. Our contributions are as follows:
\begin{itemize}[itemsep=0pt, parsep=0pt, leftmargin=10pt]
\item We propose a \textbf{unified} and \textbf{versatile} framework for generative domain adaptation, which enables multi-modal references and hybrid target domain, \eg, text-text, image-image, and image-text. We demonstrate successful adaptation to diverse domains for various generators and illustrate its advantage over other methods.
\item We demonstrate strong compositional capabilities of direction vectors in CLIP's embedding space. Taking advantage of it, we propose to linearly interpolate the direction vectors for multi-modal hybrid domain adaptation.
\item We propose a cross-domain spatial structure loss to maintain consistency with source domain. It is conducted in generator-agnostic embedding space which is versatile for various generators, \eg, StyleGAN, EG3D, and Diffusion models. To our knowledge,  it is the very first trial in generative domain adaptation.
% \item We propose a cross-domain spatial structure loss to maintain fine-grained spatial structure between source and target generator, which significantly improves consistency. It is conducted in generator-agnostic embedding space, which is the very first trial in generative domain adaptation to our knowledge.
% \item We demonstrate successful multi-modal hybrid domain adaptation to diverse new domains for various generators and illustrate its advantage over other methods. 
\end{itemize}
\section{Related Work}
\label{related}
\noindent
\textbf{Text-driven Generative Domain Adaptation.}
% 相对expansion的优势：快，无需ffhq，无需latent空间对generator无要求
Text-driven domain adaptation~\cite{gal2021stylegan, YotamNitzan2023DomainEO, Liu_2023_ICCV, song2022diffusion, lei2023diffusiongan3d, kim2023datid, kim2023podia, alanov2022hyperdomainnet, zhu2022one, lyu2023deltaedit} involves using a textual prompt to shift the domain of a pre-trained model toward a new domain. For example, Style-NADA~\cite{gal2021stylegan} presents a local direction CLIP~\cite{radford2021learning} loss to align the embeddings of the generated images and text. Based on Style-NADA, Domain Expansion (DE)~\cite{nitzan2023domain} proposes to expand the generator to jointly model multiple domains with texts.

% Although text-guided models have the potential for hybrid domain adaptation, they often have a more limited scope than image-driven domain adaptation, particularly when the domain's attributes cannot be explicitly described in a few words.
\noindent
\textbf{Image-driven Generative Domain Adaptation.}
Image-driven generative domain adaptation~\cite{mo2020freezeD, li2020fig_EWC, ojha2021fig_cdc, zhao2022closer, xiao2022few, mondalfew, wu2023domain, zhao2022few, zhao2023exploring, zhang2022towards, zhu2021mind, Alanov_2023_ICCV, duan2023weditgan,ye2023ip,kim2022diffface} refers to the adaptation of a pre-trained image generator to a new target domain using a limited number of training images. Due to the scarcity of training images, prior researches often integrate additional regularization terms to prevent overfitting. For instance, CDC~\cite{ojha2021fig_cdc} introduces the instance distance consistency loss to maintain the distance between different instances in the source domain. DiFa~\cite{zhang2022towards} utilizes GAN inversion~\cite{tov2021designing} to align the latent codes which helps inherit diversity from the source generator. Although these works have made significant strides in generative domain adaptation, they heavily rely on the discriminator or generator, making it challenging to handle hybrid domain adaptation and extend to other generators.
% RSSA~\cite{xiao2022few} proposes a relaxed spatial structural alignment method to preserve the spatial structural information of the source domain.
% AdAM~\cite{zhao2022few} introduces Adaptation-Aware Kernel Modulation to address the general few-shot image generation of various source-target domain proximities.

\noindent
\textbf{Generative Hybrid Domain Adaptation.}
To achieve hybrid domain adaptation, several domain adaptation methods propose to train a separate generative model per domain and combine their effects in test-time, \eg, Style-NADA~\cite{gal2021stylegan}. Although having potential for hybrid domain adaptation, it doubles the model size and training time due to the requirement for training multiple models separately. Domain Expansion (DE) ~\cite{nitzan2023domain} proposes to expand the generator to jointly model multiple domains via decompose latent space. However, it requires the source dataset (\eg, FFHQ~\cite{karras2019style}) for regularization, which significantly increases training time. Furthermore, it relies on the semantic latent space of the generator (\eg, StyleGAN~\cite{karras2019style} and DiffAE~\cite{preechakul2022diffusion}) for hybrid domain adaptation, limiting its applicability to a broader range of generators. Recently, FHDA~\cite{li2023fhda} proposes few-shot hybrid domain adaptation and introduces a directional subspace loss to achieve it. Differently, we focus on multi-modal references with the text and one-shot image, which offers greater flexibility and versatility.

\noindent
\textbf{Disentanglement in Generative Models.}
\quad As observed in StyleGAN~\cite{karras2019style}, the latent space is essentially a linear subspace. Recent works~\cite{harkonen2020ganspace, shen2021closed, xu2022mind, shen2020interfacegan, wu2020stylespace, patashnik2021styleclip, voynov2020unsupervised, spingarn2020gan} propose to find individual latent factors for image variations. Among them, SeFa~\cite{shen2021closed} computes the eigenvalues of the transformation matrix to find the latent directions.
% StyleAlign~\cite{wu2021stylealign} further analyzes the property of GAN's latent space and finds that the latent directions control similar semantic factors for two aligned models even though they work on far different domains.
As for diffusion models, DiffAE~\cite{preechakul2022diffusion} explores the possibility of using DPMs for representation learning and seeks to extract a meaningful and decodable representation of an input image via autoencoding. 
Drawing inspiration from the disentanglement in the latent space of them, we demonstrate that a semantically meaningful linear interpolation between the direction vectors in CLIP’s embedding space can similarly uncover favorable compositional capabilities, which facilitates us to achieve hybrid domain adaptation.

% \subsection{Self Surpervied Models}
\section{Method}
\label{sec:method}

\subsection{Multi-Modal Hybrid Domain Adaptation}
\quad We start with a pre-trained generator $G_\mathcal{S}$ (\eg, StyleGAN~\cite{karras2019style, karras2020analyzing, karras2021alias} and Diffusion model~\cite{Ho_Jain_Abbeel_2020, Song_Meng_Ermon_2020})), that maps from noise $z$ to images in a source domain $\mathcal{S}$. Given a new target domain $\mathcal{T}$ referenced by texts~\cite{gal2021stylegan, zhang2022towards, kwon2022diffusion, kim2022diffusionclip} or images~\cite{mo2020freezeD, li2020fig_EWC, ojha2021fig_cdc, zhao2022closer, xiao2022few, mondalfew}, generative domain adaptation aims to adapt $G_\mathcal{S}$ to yield a target generator $G_{\mathcal{T}}$, which can generate images similar to domain $\mathcal{T}$. 

Despite the promising results of existing methods, a major limitation of them is that they only support adaptation from the source domain to individual target domains and fail to directly adapt the generator to the hybrid domain which blends the characteristics of multiple domains. Furthermore, they fail with multi-modal adaptation driven by texts and images simultaneously. 

\begin{figure*}[t]
\centering
\includegraphics[width=0.95\linewidth]{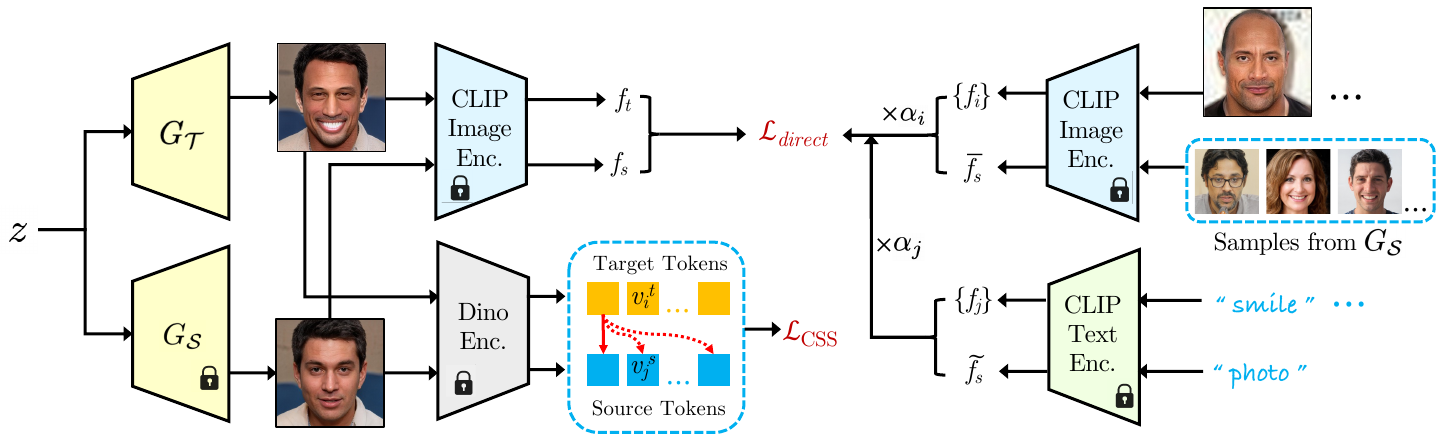}
% \vspace{-0.2cm}
\caption{Overview of UniHDA with multi-modal direction loss $\mathcal{L}_{direct}$ and cross-domain spatial structure loss $\mathcal{L}_{\text{CSS}}$. Utilizing CLIP image encoder and text encoder, $\mathcal{L}_{direct}$ encourages $G_{\mathcal{T}}$ to faithfully acquire domain-specific characteristics with multi-modal references. To facilitate diversity inherited from $G_{\mathcal{S}}$, $\mathcal{L}_{\text{CSS}}$ improves cross-domain consistency by maintaining detailed spatial structure information. The red solid line represents positive pairs, while red dashed lines represent negative pairs.}
\label{fig:method}
\vspace{-0.3cm}
\end{figure*}

For more general purposes, we propose multi-modal hybrid domain adaptation. Given $N$ domains $\{\mathcal{T}_i\}_{i=1}^N$ with one-shot image $\{Y_i\}$ and $M$ domains $\{\mathcal{T}_j\}_{j=1}^M$ with the text prompt $\{P_j\}$, it aims to adapt the source generator $G_{\mathcal{S}}$ to $G_{\mathcal{T}}$ that models the hybrid domain $\mathcal{T}=\{\mathcal{T}_i\}\cup\{\mathcal{T}_j\}$ and generates images with integrated characteristics. To the end, we introduce UniHDA, a unified and versatile framework for multi-modal hybrid domain adaptation (\cref{fig:method}).

\subsection{Multi-modal Direction Loss}
\quad To enable multiple modalities, we leverage pre-trained CLIP~\cite{radford2021learning} to encode text-image references into a unified semantic embedding space. Drawing inspiration from CLIP-based methods~\cite{gal2021stylegan, zhang2022towards, kwon2022diffusion, kim2022diffusionclip}, we represent the \textit{domain shift} as the direction vector $\Delta f_{dom}$ from the source embedding to the target embedding. For image reference $Y_i$ and its CLIP embedding $f_i$, the \textit{domain shift} is calculated by
\begin{equation}
\begin{gathered}
\Delta f_{dom} = f_i - \overline{f_s},
\end{gathered}
\end{equation}
where $\overline{f_s}$ is the mean embedding of several samples generated by $G_\mathcal{S}$.
For text prompt $P_j$ and its CLIP embedding $f_j$,  
\begin{equation}
\begin{gathered}
\Delta f_{dom} = f_j - \widetilde{f_s},
\end{gathered}
\end{equation}
where $\widetilde{f_s}$ is the embedding of the source text prompt.

To adapt $G_{\mathcal{S}}$, we initialize a new generator $G_{\mathcal{T}}$ from $G_{\mathcal{S}}$ and finetune it by aligning the \textit{sample-shift} direction $\Delta f_{samp}$ with the \textit{domain-shift} direction $\Delta f_{dom}$. Formally, 
\begin{equation}
\label{eq:direct}
\begin{gathered}
\Delta f_{samp} = f_t - f_s, \\
\mathcal{L}_{direct} = 1 - \frac{\Delta f_{samp} \cdot \Delta f_{dom}}{\left\|\Delta f_{samp}\right\|\left\|\Delta f_{dom}\right\|},
\end{gathered}
\end{equation}
where $f_s$ and $f_t$ are the embeddings of samples generated by $G_\mathcal{S}$ and $G_\mathcal{T}$ with the same noise.

\subsection{Linear Composition of Direction Vectors}
\quad To achieve the hybrid domain adaptation, we draw inspiration from the compositional capabilities in the latent space of StyleGAN~\cite{harkonen2020ganspace, shen2021closed, xu2022mind}. We illustrate that a linear interpolation between two direction vectors in the embedding space of CLIP, which is semantically meaningful, reveals promising compositional capabilities. As shown in \cref{fig:linear}, we can smoothly interpolate between two direction vectors calculated by distinct target prompts and source prompt ``\textit{photo}'', resulting in a gradual adaptation toward the target domain.

In light of this intriguing finding, we employ linear interpolation on the direction vectors of multi-modal target domains, to derive a consolidated direction vector representing the hybrid domain that semantically integrates all attributes. For given domain coefficients $\{\alpha_i\}$ and $\{\alpha_j\}$, we obtain the direction vector 
\begin{equation}
\label{eq:linear}
\begin{gathered}
\Delta f_{dom} = \sum_{i=1}^N\alpha_i(f_i - \overline{f_s}) + \sum_{j=1}^M\alpha_j(f_j - \widetilde{f_s}),
\end{gathered}
\end{equation}
which represents the \textit{domain shift} between the hybrid domain and source domain. We then substitute \cref{eq:linear} into \cref{eq:direct} to adapt $G_\mathcal{S}$ to the hybrid domain. 

\subsection{Cross-domain Spatial Structure Loss}
\quad Albeit the direction loss achieves multi-modal hybrid domain adaptation, the adapted generator are prone to overfit domain-specific attributes. This exacerbates when it comes to image-image and image-text scenarios owing to the scarcity of the images. To address this issue, we introduce a novel cross-domain spatial structure loss (CSS) to enhance cross-domain consistency, ensuring the preservation of intricate spatial structural information between the source and target generator. 

Specifically, we leverage the pre-trained Dino-ViT~\cite{dosovitskiy2020image, oquab2023dinov2} to encode the generated images into patch tokens, containing detailed spatial structural information. Dino-ViT is self-supervised trained to focus on the distinction between subjects of the same class~\cite{ruiz2023dreambooth}, which facilitates us to maintain the cross-domain consistency. Motivated by contrastive learning~\cite{oord2018representation}, we reduce the distance between the positive token pairs at the same position and push away the negative token pairs at the other positions by 
\begin{equation}
\begin{gathered}
\mathcal{L}_{\text{CSS}} = -\sum_{i}\log\frac{\exp(v_i^t\cdot v_i^s)}{\sum_{j}\exp(v_i^t\cdot v_j^s)},
\end{gathered}
\end{equation}
where $v_i^t$ and $v_j^s$ are the $i$-th and $j$-th tokens in the last layer of Dino-ViT from $G_\mathcal{T}$ and $G_\mathcal{S}$ respectively. The dot mark $\cdot$ represents dot product.

Overall, our training loss consists of two terms, i.e., the multi-modal direction loss $\mathcal{L}_{direct}$ to achieve multi-modal hybrid domain adaptation and the cross-domain spatial structure loss $\mathcal{L}_{\text{CSS}}$ to maintain cross-domain consistency:
\begin{equation}
\begin{gathered}
\mathcal{L}_{overall} = \mathcal{L}_{direct} + \lambda \mathcal{L}_{\text{CSS}},
\end{gathered}
\end{equation}
where we use $\lambda=5$ in our experiments.

\section{Experiments}
\subsection{Experimental Setting}
\textbf{Methodology.}
We demonstrate the versatility of UniHDA on multi-modal hybrid domain adaptation, \ie, image-image, text-text, and image-text. To show the generator-agnostic nature of UniHDA, we apply it to three well-known generators, \ie,  StyleGAN2~\cite{karras2020analyzing}, Diffusion model~\cite{kim2022diffusionclip}, and EG3D~\cite{chan2022efficient}. Following previous generative domain adaptation literatures ~\cite{gal2021stylegan, zhang2022towards,mo2020freezeD, li2020fig_EWC, nitzan2023domain, ojha2021fig_cdc, zhao2022closer, xiao2022few, mondalfew}, we use StyleGAN2 for comparisons in most experiments. 

% \noindent
% \textbf{Baselines.}
% To demonstrate the effectiveness of our method, we compare our method against diverse baselines methods. For text-text references, we compare it with 

\noindent
\textbf{Datasets.}
We conduct the experiments for a wide range of source and target domains to validate the effectiveness of UniHDA. Following previous work, we consider FFHQ \cite{karras2019style}, AFHQ-Dog \cite{choi2020stargan}, and LSUN-Church \cite{yu2015lsun} as the source domains. The resolutions of images in these datasets are respectively 1024, 512, and 256. We adapt the pre-trained models to diverse hybrid domains driven by the text prompt and one-shot image. To demonstrate the effect of the hybrid domain, \textit{we set the domain coefficients in \cref{eq:linear} as 0.5.}

\begin{figure*}[t]
\centering
\includegraphics[width=.95\linewidth]{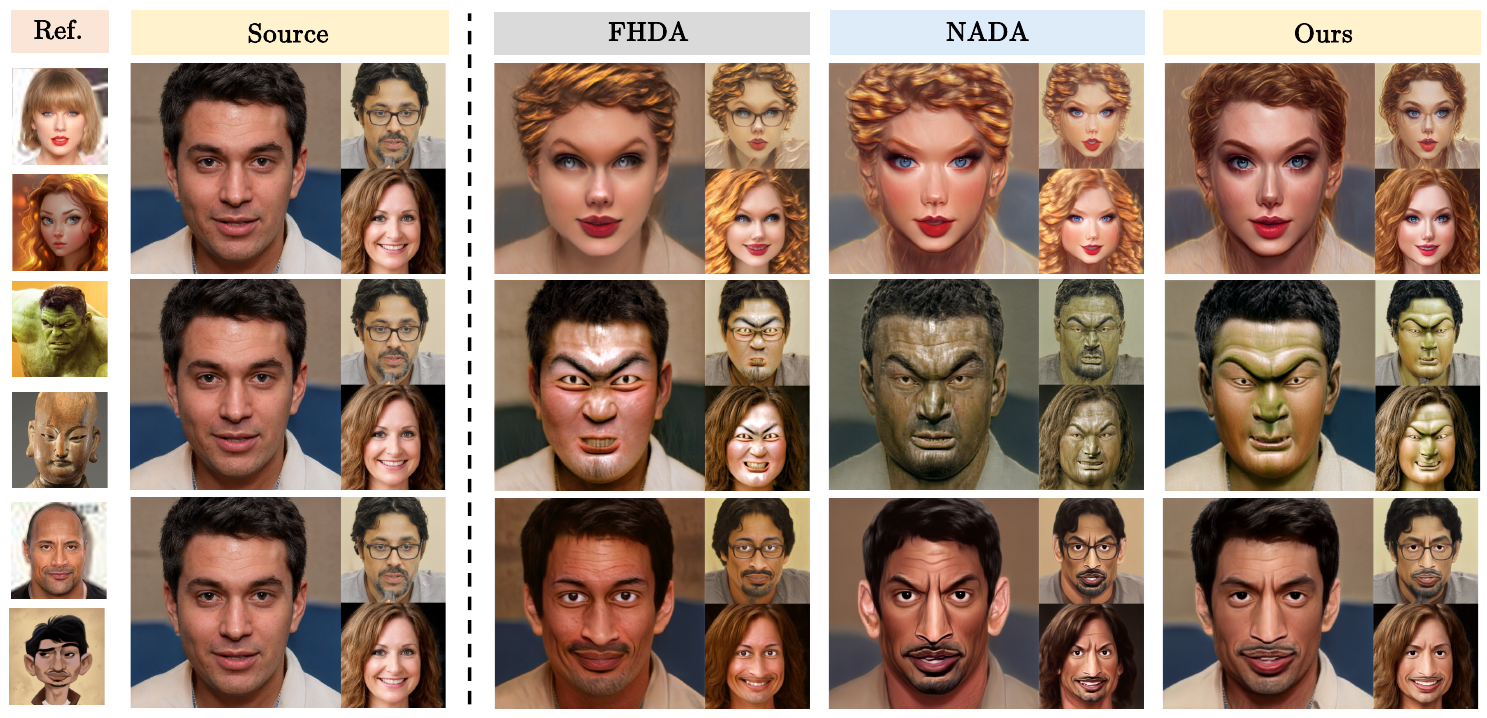}
\vspace{-0.2cm}
\caption{\textbf{Image-image} hybrid domain adaptation. We compare the results of FHDA~\cite{li2023fhda}, NADA~\cite{gal2021stylegan} and UniHDA (Ours) with the same noise. FHDA and NADA generate images with poor cross-domain consistency, leading to a limited diversity. In contrast, UniHDA alleviates overfitting and maintains strong cross-domain consistency.}
\label{fig:img}
\vspace{-0.2cm}
\end{figure*}

\begin{table*}[t]
\small
% \centering
\begin{center}
\setlength{\tabcolsep}{4pt}
\renewcommand\arraystretch{1.1}
\begin{tabular}{ccccccccc}
\toprule
\multicolumn{1}{c}{\multirow{2}{*}{Method}}& \multicolumn{2}{c}{\textit{Taylor-Elena}} & \multicolumn{2}{c}{\textit{Hulk-Wooden}} &
\multicolumn{2}{c}{\textit{Johnson-Comic}}  
&\multicolumn{2}{c}{Average}
\\
\cmidrule(lr){2-3} \cmidrule(lr){4-5} \cmidrule(lr){6-7} \cmidrule(lr){8-9}
    \multicolumn{1}{c}{}
    & {\scriptsize{CS-I} ($\uparrow$)}
    & {\scriptsize{SCS} ($\uparrow$)} 
    & {\scriptsize{CS-I} ($\uparrow$)}   
    & {\scriptsize{SCS} ($\uparrow$)}
    & {\scriptsize{CS-I} ($\uparrow$)} 
    & {\scriptsize{SCS} ($\uparrow$)}
    & {\scriptsize{CS-I} ($\uparrow$)} 
    & {\scriptsize{SCS} ($\uparrow$)}
    \\ \midrule 
FHDA
&0.685 &0.576
&0.635 &0.659
&0.640 &0.679
&0.630 &0.661
 \\ 
 NADA 
&0.684	&0.579
&0.624	&0.575
&0.647	&0.642
&0.628 &0.639
 \\ 
 Ours   
&\textbf{0.699}	&\textbf{0.738}
&\textbf{0.649}	&\textbf{0.707}
&\textbf{0.656}	&\textbf{0.764}
&\textbf{0.642} &\textbf{0.769}
 \\ 
 \bottomrule

\end{tabular}
% \centering
\end{center}
\vspace{-0.2cm}
\caption{Quantitative results for \textbf{image-image} domain adaptation. We present the quantitative results corresponding to each case in \cref{fig:img}. To further demonstrate the robustness of our method, we average the results for 25 cases (shown in Appendix).}
\label{tab:img}
\vspace{-0.5cm}
\end{table*}

\noindent
\textbf{Evaluation Metrics.}
One important aspect to evaluate generative domain adaptation is the preservation of domain-specific characteristics. Following ~\cite{ruiz2023dreambooth}, we use CLIP Score (CS-T and CS-I) for text-text and image-image adaptation respectively. Concretely, CS-T is measured by the average cosine similarity between prompt and generated images' embedding. CS-I is the average pairwise cosine similarity between CLIP embeddings of generated and real images. Here we use average CS-T or CS-I of multiple domains. For image-text adaptation, we use the average of CS-T and CS-I as the metric (CS). 

Another important evaluation is the cross-domain consistency of the source domain. To measure it, we adopt Structural Consistency Score (SCS)~\cite{xiao2022few} to evaluate the spatial structural consistency between source and target generator.

\begin{figure*}[t]
\centering
\includegraphics[width=.95\linewidth]{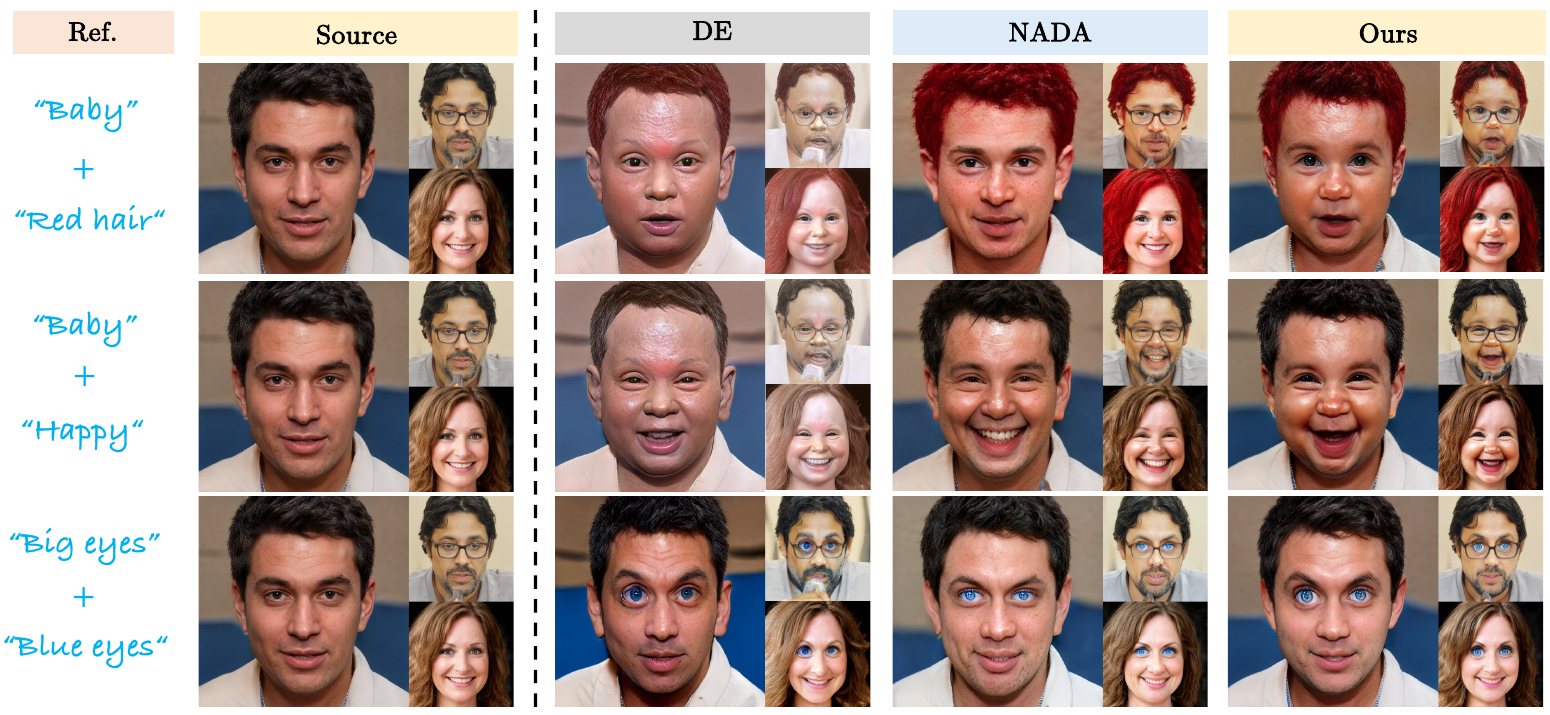}
\vspace{-0.2cm}
\caption{\textbf{Text-text} hybrid domain adaptation. We compare the results of DE~\cite{nitzan2023domain}, NADA~\cite{gal2021stylegan} and UniHDA (Ours) with the same noise. UniHDA exhibits desirable performance to acquire characteristics from hybrid target domain and maintain robust cross-domain consistency.}
\label{fig:text}
\vspace{-0.2cm}
\end{figure*}

\subsection{Image-image Hybrid Domain Adaptation}
\label{img}
% baseline: nada(model interpolation) HDA
% \textbf{Qualitative Results.}
\quad \cref{fig:img} shows the qualitative results of baselines and UniHDA for image-image adaptation, starting from the same source domain FFHQ~\cite{karras2019style} to the combinations of individual domains. As shown in the figure, FHDA~\cite{li2023fhda} suffers from severe model collapse and generates images with a limited diversity due to the scarcity of image references. While NADA~\cite{gal2021stylegan} mitigates overfitting to a certain extent, its cross-domain consistency remains poor, resulting in the generation of similar images. In contrast, UniHDA maintains strong consistency and effectively generates images with characteristics of the hybrid domain.

% \noindent
% \textbf{Quantitative Results.}
We also quantitatively compare UniHDA with baselines. As shown in \cref{tab:img}, ours clearly outperforms them. 
For CS-I, UniHDA significantly outperforms other methods, indicating that generated images effectively integrate multiple characteristics from distinct domains. Furthermore, UniHDA achieves better SCS, which effectively maintains cross-domain consistency compared with baselines.

\begin{table*}[t]
\small
% \centering
\begin{center}
\setlength{\tabcolsep}{4pt}
\renewcommand\arraystretch{1.1}
\begin{tabular}{ccccccccc}
\toprule
\multicolumn{1}{c}{\multirow{2}{*}{Method}}& \multicolumn{2}{c}{\textit{Baby-Red hair}} & \multicolumn{2}{c}{\textit{Baby-Happy}} &
\multicolumn{2}{c}{\textit{Big-Blue eyes}}  
&\multicolumn{2}{c}{Average}
\\
\cmidrule(lr){2-3} \cmidrule(lr){4-5} \cmidrule(lr){6-7}
\cmidrule(lr){8-9}    \multicolumn{1}{c}{}
    & {\scriptsize{CS-T} ($\uparrow$)}
    & {\scriptsize{SCS} ($\uparrow$)} 
    & {\scriptsize{CS-T} ($\uparrow$)}   
    & {\scriptsize{SCS} ($\uparrow$)}
    & {\scriptsize{CS-T} ($\uparrow$)} 
    & {\scriptsize{SCS} ($\uparrow$)}
    & {\scriptsize{CS-T} ($\uparrow$)} 
    & {\scriptsize{SCS} ($\uparrow$)}
    \\ \midrule 
DE  
&0.163  &0.638
&0.160  &0.580
&0.195  &0.662
&0.167  &0.634
\\ 
NADA 
&0.179	&0.661
&0.170	&0.642
&0.186	&0.731
&0.159  &0.552
\\ 
Ours   
&\textbf{0.186}	&\textbf{0.744}
&\textbf{0.175}	&\textbf{0.757}
&\textbf{0.197}	&\textbf{0.765}
&\textbf{0.176} &\textbf{0.707}
 \\ 
 \bottomrule

\end{tabular}
% \centering
\end{center}
\vspace{-0.2cm}
\caption{Quantitative results for \textbf{text-text} domain adaptation. We present the quantitative results corresponding to each case in \cref{fig:text}. Similar to \cref{tab:img}, we average the quantitative results for 25 cases (shown in Appendix).}
\label{tab:text}
\vspace{-0.3cm}
\end{table*}

\begin{figure*}[t]
\centering
\includegraphics[width=.98\linewidth]{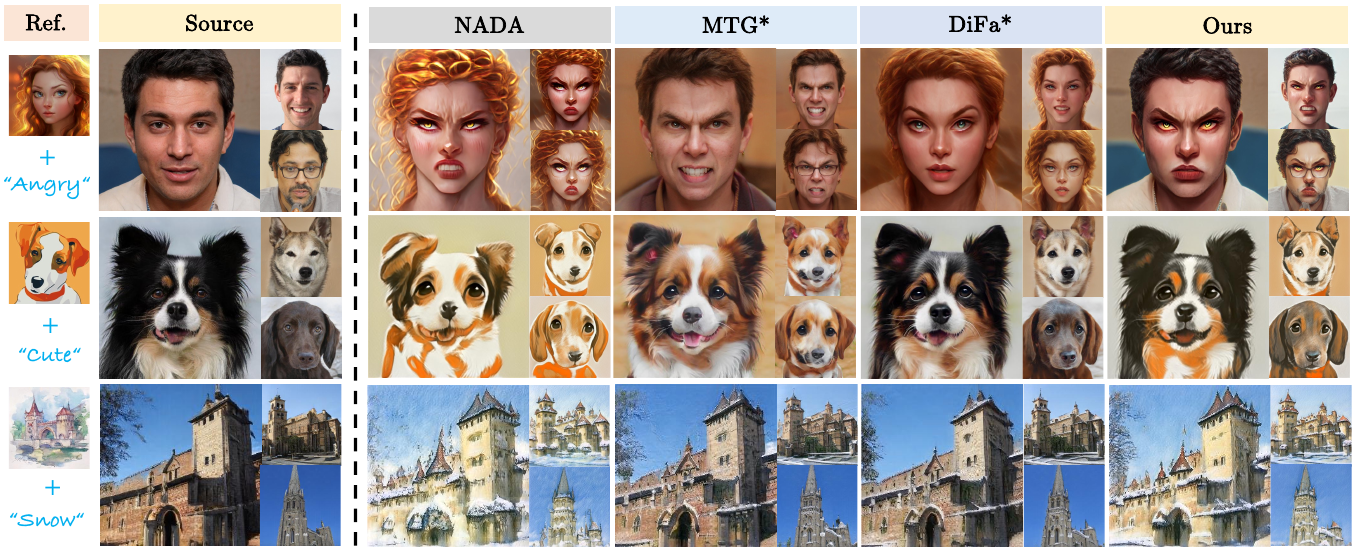}
\vspace{-0.2cm}
\caption{\textbf{Image-text} hybrid domain adaptation. We compared our method with previous~\cite{gal2021stylegan, zhu2021mind, zhang2022towards}, UniHDA well captures the attributes of hybrid target domain and maintains strong cross-domain consistency with source domain. $*$ indicates that MTG and DiFa support multi-modalities by interpolating model parameters with NADA.}
\label{fig:img-text}
\vspace{-0.2cm}
\end{figure*}

\subsection{Text-text Hybrid Domain Adaptation}
\label{text}
% \textbf{Qualitative Results.}
\quad \cref{fig:text} shows the qualitative results for text-text adaptation. Since the adaptation is conducted solely along one projection direction of the latent code, Domain Expansion (DE)~\cite{nitzan2023domain}, to some extent, does not fully capture the characteristics of the target domain, \eg, \textit{baby} (Row 1 and Row 2). Furthermore, DE does not maintain robust consistency, \eg, the chin of the person in the upper-right corner and background artifacts in Row 3. This is primarily due to the characteristics bias introduced when projecting the latent code to the base subspace, making the images less authentic. The problem of NADA~\cite{gal2021stylegan} is overfitting. Hard-to-learn characteristics, \eg, \textit{baby} (Row 1 and Row 2) and \textit{big eyes} (Row 3) may be overshadowed by other overfitted ones. In contrast, UniHDA (Ours) exhibits desirable performance to generate images with integrated characteristics while maintaining robust consistency with the source domain. 

Similar to \cref{img}, we also compare UniHDA with the baselines quantitatively. As shown in \cref{tab:text}, ours clearly outperforms the baselines, which are consistent with qualitative results in \cref{fig:text}. We achieve better CS-I and SCS, indicating that generated images effectively integrate domain-specific attributes and preserve primary characteristics of the source domain.

\begin{table}[t]
\small
\setlength{\tabcolsep}{0.4pt}
\renewcommand\arraystretch{1.0}
\begin{minipage}{0.58\textwidth}
% \centering
\begin{center}
\begin{tabular}{ccccccc}
\toprule
\multicolumn{1}{c}{\multirow{2}{*}{Method}}& 
\multicolumn{2}{c}{FFHQ }  &
\multicolumn{2}{c}{Dog }  &
\multicolumn{2}{c}{Church } 
\\
\cmidrule(lr){2-3} \cmidrule(lr){4-5} \cmidrule(lr){6-7} 
    \multicolumn{1}{c}{}
    & {\scriptsize{CS} ($\uparrow$)}
    & {\scriptsize{SCS} ($\uparrow$)} 
    & {\scriptsize{CS} ($\uparrow$)}
    & {\scriptsize{SCS} ($\uparrow$)} 
    & {\scriptsize{CS} ($\uparrow$)}
    & {\scriptsize{SCS} ($\uparrow$)} 
    \\ \midrule 
 \scriptsize{NADA} 
&\scriptsize{0.563} &\scriptsize{0.586}	
&\scriptsize{0.424} &\scriptsize{0.533}  
&\scriptsize{\textbf{0.414}} &\scriptsize{0.629}  
 \\ 
 \scriptsize{MTG}
&\scriptsize{0.536} &\scriptsize{0.529} 
&\scriptsize{0.403} &\scriptsize{0.526} 
&\scriptsize{0.403} &\scriptsize{0.684} 
 \\ 
\scriptsize{DiFa} 
&\scriptsize{0.548}	&\scriptsize{0.681} 
&\scriptsize{0.413}	&\scriptsize{0.683} 
&\scriptsize{0.407}	&\scriptsize{0.711} 
\\
 \scriptsize{Ours  } 
&\scriptsize{\textbf{0.565}}&\scriptsize{\textbf{0.742}}
&\scriptsize{\textbf{0.430}}&\scriptsize{\textbf{0.796}}
&\scriptsize{\textbf{0.414}}&\scriptsize{\textbf{0.781}}
 \\ 
 \bottomrule
\end{tabular}
\end{center}
\vspace{-0.2cm}
\caption{Quantitative results for \textbf{image-text} adaptation. We average the quantitative results for 81 cases for FFHQ, 16 cases for AFHQ-Dog, and 16 cases for LSUN-Church (shown in Appendix).}
\label{tab:img-text}
\vspace{-0.8cm}
\end{minipage}
\hspace{0.1cm}
\begin{minipage}{0.4\textwidth}
\small
\begin{center}
\setlength{\tabcolsep}{0.4pt}
\renewcommand\arraystretch{1.0}
\begin{tabular}{cccc}
\toprule
Method & Fidel. & Diver. & Corr.
\\
 \midrule 
\scriptsize{vs. NADA (I-I)}
&\scriptsize{85.2} &\scriptsize{90.6}  & \scriptsize{76.0}  	
 \\ 
\scriptsize{vs. NADA (T-T)}
&\scriptsize{81.4} &\scriptsize{84.2}  &\scriptsize{80.8}  
\\
\scriptsize{vs. NADA (T-I)}
&\scriptsize{84.6} &\scriptsize{85.8}	&\scriptsize{78.6}
 \\ 
 \bottomrule

\end{tabular}
% \centering
% \end{center}
% \centering
\end{center}
\vspace{-0.2cm}
\caption{User study for fidelity, diversity, and reference corespondence (image or text) in hybrid domain adaptation. The value (\%) represents the percentage of users who favor the images generated by our method over NADA. }
\label{tab:user}
\vspace{-0.8cm}
\end{minipage}
\end{table}

% \begin{table}[t]
% \small
% \begin{center}
% \setlength{\tabcolsep}{4pt}
% \renewcommand\arraystretch{1.1}
% \begin{tabular}{cccc}
% \toprule
% Method & Fidelity & Diversity & Corr.
% \\
%  \midrule 
% vs. NADA (I-I)
% &85.2\% &90.6\%  & 76.0\%  	
%  \\ 
% vs. NADA (T-T)
% &81.4\% &84.2\%  &80.8\%  
% \\
% vs. NADA (T-I)
% &84.6\% &85.8\%	&78.6\%
%  \\ 
%  \bottomrule

% \end{tabular}
% \vspace{-0.2cm}
% % \centering
% \end{center}
% \caption{User study. Corr. represents image or text correspondence. The value represents the percentage of users who favor the images generated by our method over the other. For each case, we generate 1000 samples and randomly assign 200 samples to 30 users. }
% \label{tab:user}
% \vspace{-0.3cm}
% \end{table}

\begin{table}[t]
\small
% \centering
\begin{center}
\setlength{\tabcolsep}{3pt}
\renewcommand\arraystretch{1.1}
% \begin{threeparttable}
\begin{tabular}{cccccccc}
\toprule
\multicolumn{1}{c}{\multirow{2}{*}{Method}}&
\multicolumn{1}{c}{\multirow{2}{*}{Modality}}&
\multicolumn{1}{c}{\multirow{2}{*}{\makecell[c]{\scriptsize{Generator} \\ \scriptsize{Dependency}}}}&
\multicolumn{1}{c}{\multirow{2}{*}{\makecell[c]{\scriptsize{Model} \\ \scriptsize{Amount}}}}&
\multicolumn{2}{c}{2-domain}& 
\multicolumn{2}{c}{10-domain}  
\\
\cmidrule(lr){5-6} \cmidrule(lr){7-8} 
    \multicolumn{1}{c}{}
    &\multicolumn{1}{c}{}
    &\multicolumn{1}{c}{}
    &\multicolumn{1}{c}{}
    & {\scriptsize{size} ($\downarrow$)}
    & {\scriptsize{time} ($\downarrow$)}    
    & {\scriptsize{size} ($\downarrow$)} 
    & {\scriptsize{time} ($\downarrow$)}\\
    \midrule 
 NADA & Multi & - 
& \scriptsize{$N$} 
&48\scriptsize{M} &4\scriptsize{min}
&240\scriptsize{M} &20\scriptsize{min}
\\
MTG$^*$ & Multi
& \scriptsize{StyleGAN} 
& \scriptsize{$N$} 
&48\scriptsize{M} &4\scriptsize{min}
&240\scriptsize{M} &20\scriptsize{min}
\\
DiFa$^*$ & Multi
& \scriptsize{StyleGAN} 
& \scriptsize{$N$} 
&48\scriptsize{M} &4\scriptsize{min}
&240\scriptsize{M} &20\scriptsize{min}
\\
DE$^\dag$ & Text
& \scriptsize{Latent space} 
&\scriptsize{1}
&24\scriptsize{M} &20\scriptsize{h}
&24\scriptsize{M} &20\scriptsize{h}  
 \\ 
FHDA & Image
& -
&\scriptsize{1}
&24\scriptsize{M} &3\scriptsize{min}
&24\scriptsize{M} &3\scriptsize{min}  
 \\ 
\midrule
Ours & Multi &- 
&\scriptsize{1} 
&\textbf{24\scriptsize{M}} &\textbf{2\scriptsize{min}}
&\textbf{24\scriptsize{M}} &\textbf{2\scriptsize{min}}

 \\ 
 \bottomrule

\end{tabular}
% \begin{tablenotes}   
% \footnotesize              
% \item $*$ indicates that DE needs a semantic latent space of the generator.
% \end{tablenotes}           
% \end{threeparttable} 
% \centering
\end{center}
\vspace{-0.2cm}
\caption{Comparison with previous methods. $*$ indicates that MTG and DiFa support multi-modalities by interpolating model parameters with NADA. $\dag$ means that DE needs source dataset (e.g., FFHQ) that significantly increases the training time. Besides, MTG, DiFa, and DE have additional dependencies on the type of generator, which limits their broader applicability.
}
\label{tab:time}%
\vspace{-0.5cm}
\end{table}

\begin{figure*}[t]
\centering
\includegraphics[width=.95\linewidth]{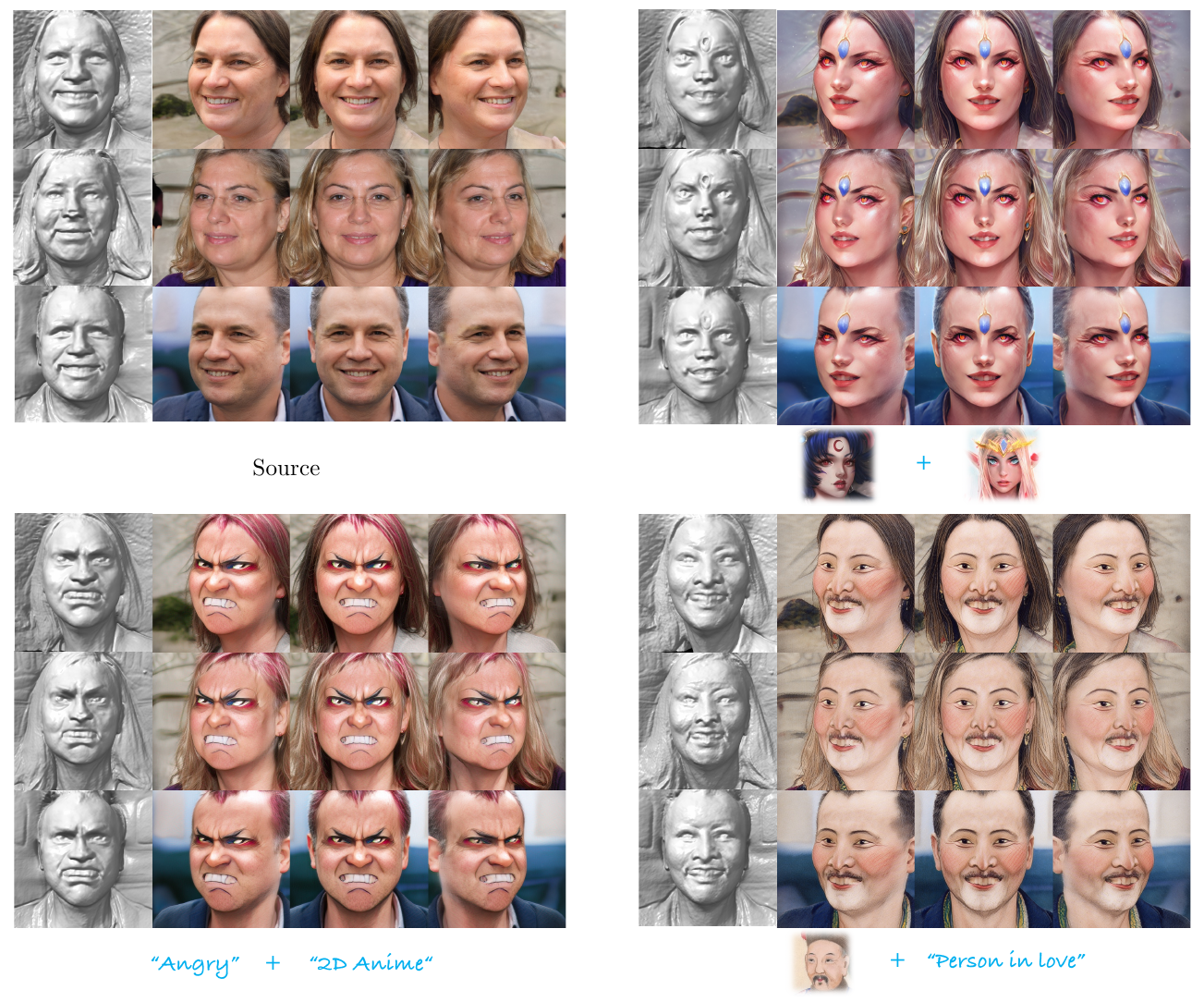}
\vspace{-0.2cm}
\caption{Hybrid domain adaptation in 3D generator. To show the versatility of UniHDA, we apply it on the popular 3D-aware generator, EG3D~\cite{chan2022efficient}.}
\label{fig:3d}
\vspace{-0.3cm}
\end{figure*}

\begin{figure*}[t]
\centering
\includegraphics[width=0.95\linewidth]{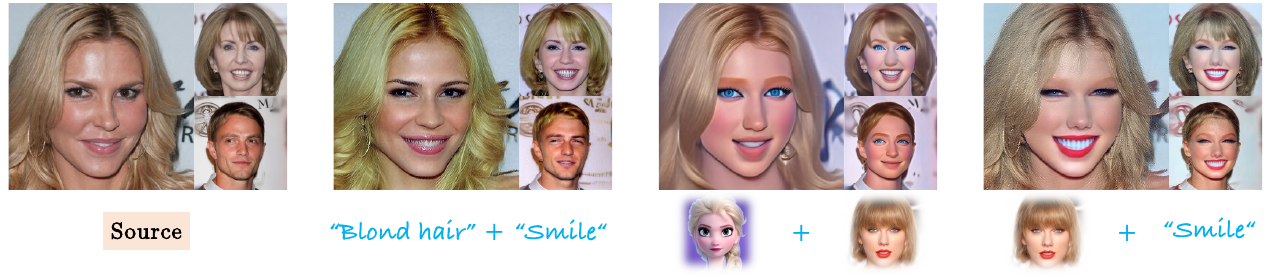}
\vspace{-0.2cm}
\caption{Results of UniHDA with DiffusionCLIP~\cite{kim2022diffusionclip}, which demonstrate UniHDA is agnostic to the type of generator, allowing for broader application on diffusion models.
% We replace the training objective of DiffusionCLIP with our proposed $\mathcal{L}_{direct}$ and $\mathcal{L}_{\text{CSS}}$.
}
\label{fig:diff}
\vspace{-0.3cm}
\end{figure*}

% ninihihu
\begin{figure*}[t]
\centering
\includegraphics[width=0.95\linewidth]{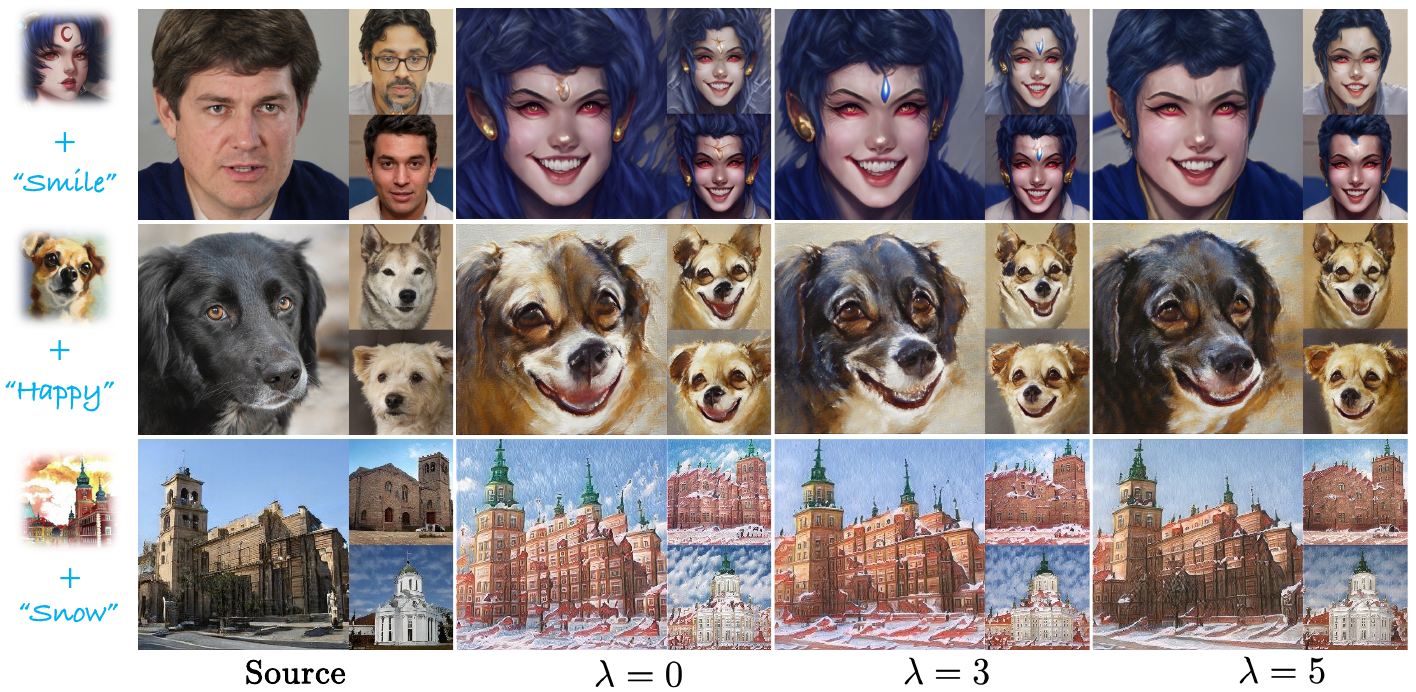}
% \vspace{-0.2cm}
\caption{Ablation of our proposed $\mathcal{L}_{\text{CSS}}$ on hybrid domain adaptation, which significantly alleviates overfitting and improves cross-domain consistency.}
\label{fig:ablation}
\vspace{-0.3cm}
\end{figure*}

\subsection{Image-text Hybrid Domain Adaptation}
\quad \cref{fig:img-text} shows the results of image-text adaptation, including FFHQ~\cite{karras2019style}, AFHQ-Dog~\cite{choi2020stargan}, and LSUN
-Church~\cite{yu2015lsun}. As depicted in \cref{img}, NADA is susceptible to overfitting, which retains poor cross-domain consistency. Besides, we interpolate NADA's parameters with MTG\cite{zhu2021mind} and DiFa~\cite{ zhang2022towards}, which alleviates overfitting to some extent. However, they can't accurately capture the attributes of hybrid target domain and still fail to maintain good consistency. In contrast, UniHDA well captures the attributes and achieves robust consistency in all scenarios.
% More results are included in Appendix.

As shown in \cref{tab:img-text}, we also compare UniHDA with the baselines quantitatively. Consistent with qualitative results in \cref{fig:img-text}, ours clearly outperforms the baselines. Besides, we conduct user study in \cref{tab:user}. For each case, we generate 1000 samples and randomly assign 200 samples to 30 users. The results indicate that UniHDA surpasses NADA in terms of fidelity, diversity and reference correspondence.

\subsection{Comparison with Existing Methods}
% img(Difa) and text(nada)
\quad In addition to surpassing baselines for generation quality, UniHDA also surpasses them in terms of efficiency, \eg, model size and training time as shown in \cref{tab:time}. NADA, MTG, and DiFa trains a separate generative model per domain and interpolates their parameters in test-time, which necessitates multiple times the model size and training time. Although DE avoids cross-model interpolation, it heavily relies on the large source dataset (\eg, FFHQ~\cite{karras2019style}) for regularization during training process, resulting in a significant increase in training time. In contrast, UniHDA circumvents these issues, which enables the completion of the adaptation within single generator in only two minutes. 

Furthermore, DE relies on the semantic latent space of the generator (\eg, StyleGAN~\cite{karras2019style} and DiffAE~\cite{preechakul2022diffusion}) for hybrid domain adaptation, limiting its applicability to a broader range of generators. MTG and DiFa utilize GAN inversion, which restricts the applicability to generators similar to StyleGAN. Conversely, UniHDA is not constrained by the type of generators, allowing for its broader application across various generators.

\begin{figure*}[t]
\centering
\includegraphics[width=.9\linewidth]{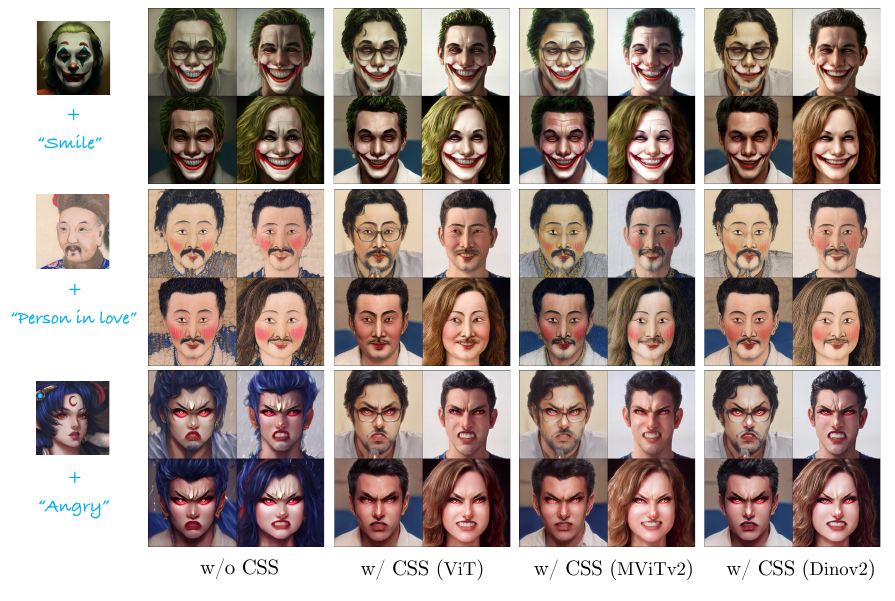}
\vspace{-0.4cm}
\caption{Ablation of different pre-trained encoders for CSS on hybrid domain adaptation.}
\label{fig:encoder-it}
\vspace{-0.3cm}
\end{figure*}

\begin{table}[t]
\small
% \centering
\begin{center}
\setlength{\tabcolsep}{2pt}
\renewcommand\arraystretch{1.1}
\begin{tabular}{ccccccccccc}
\toprule
\multicolumn{1}{c}{\multirow{2}{*}{$\lambda$}}& \multicolumn{2}{c}{FFHQ \scriptsize{(I-I)}} & \multicolumn{2}{c}{FFHQ \scriptsize{(T-T)}} &
\multicolumn{2}{c}{FFHQ \scriptsize{(T-I)}}  &
\multicolumn{2}{c}{Dog \scriptsize{(T-I)}}  &
\multicolumn{2}{c}{Church \scriptsize{(T-I)}} 
\\
\cmidrule(lr){2-3} \cmidrule(lr){4-5} \cmidrule(lr){6-7} \cmidrule(lr){8-9} \cmidrule(lr){10-11}
\multicolumn{1}{c}{}
& {\scriptsize{SCS} ($\uparrow$)} 
& {\scriptsize{CS-I} ($\uparrow$)} 
& {\scriptsize{SCS} ($\uparrow$)} 
& {\scriptsize{CS-T} ($\uparrow$)}
& {\scriptsize{SCS} ($\uparrow$)} 
& {\scriptsize{CS} ($\uparrow$)}
& {\scriptsize{SCS} ($\uparrow$)} 
& {\scriptsize{CS} ($\uparrow$)}
& {\scriptsize{SCS} ($\uparrow$)} 
& {\scriptsize{CS} ($\uparrow$)}
\\ \midrule 
0 
&0.502 &0.639
&0.520 &0.170
&0.562 &0.557 
&0.491 &\textbf{0.430} 
&0.604 &0.411 
 \\ 
3 
&0.681 &0.638	
&0.683 &0.171	
&0.694 &0.556	
&0.787 &0.428
&0.706 &0.413
\\
5  
&\textbf{0.769}	&\textbf{0.642}
&\textbf{0.707}	&\textbf{0.176}
&\textbf{0.742}&\textbf{0.565}
&\textbf{0.796}&\textbf{0.430}
&\textbf{0.781}&\textbf{0.414}
% \\
% 10
% &\textbf{0.742}&\textbf{0.565}
% &\textbf{0.796}&\textbf{0.430}
% &\textbf{0.781}&\textbf{0.414}
 \\ 
 \bottomrule
\end{tabular}
% \centering
\end{center}
\vspace{-0.2cm}
\caption{Quantitative ablation for our proposed $\mathcal{L}_{\text{CSS}}$.}
\label{tab:ablation}
\vspace{-0.8cm}
\end{table}

\subsection{Generalization on 3D Generator}
\quad To further verify the versatility of UniHDA, we apply it on the popular 3D-aware image generation method, EG3D~\cite{chan2022efficient}. Specifically, we replace the discrimination loss with our proposed $\mathcal{L}_{direct}$ and $\mathcal{L}_{\text{CSS}}$ for hybrid domain adaptation. As shown in \cref{fig:3d}, the results effectively integrate the attributes and preserve the characters and poses of source domain, demonstrating that UniHDA is agnostic to the type of generative models and can be easily extended to other generators.

\subsection{Generalization on Diffusion Model}
\label{sec:diff}
\quad In this section, we demonstrate that UniHDA is agnostic to the type of generative models and can easily generalize to diffusion models. Specifically, we apply UniHDA on DiffusionCLIP~\cite{kim2022diffusionclip} and replace the training objective of DiffusionCLIP with our proposed $\mathcal{L}_{direct}$ and $\mathcal{L}_{\text{CSS}}$. As shown in \cref{fig:diff}, the results integrate the characteristics from multiple target domains and maintain robust consistency with the source domain. This verifies desirable generalization ability and versatility of UniHDA. More results are included in Appendix.

% \subsection{traversal}
\subsection{Ablation of CSS Loss}
\label{ablation}
% ablation of dino loss
\quad We conduct the ablation study to evaluate the effects of our proposed Cross-domain Spatial Structure (CSS) loss on single domains. As shown in \cref{fig:ablation}, the results without $\mathcal{L}_{\text{CSS}}$ suffer from overfitting and have very limited cross-domain consistency, \eg, distorted backgrounds in Row 1 and 3. Benefited from CSS, the generated images maintain consistency with the source images in terms of spatial structure, thereby inheriting the diversity from the source domain. More results are included in Appendix. Besides, we conduct the quantitative ablation in \cref{tab:ablation}, which is consistent with the qualitative results. 

\subsection{Ablation of Encoder for CSS}
\quad We conduct experiments on pre-trained ViT~\cite{dosovitskiy2020image}, MViTv2~\cite{li2022mvitv2}, and Dinov2 to explore the impact of different image encoders for CSS. As shown in \cref{fig:encoder-it}, we can observe that all of them improve the consistency with source domain compared with the baseline approach. Furthermore, they exhibit a similar qualitative style, which demonstrates that our CSS is agnostic to different pre-trained image encoders. More results are included in Appendix.
% \vspace{-0.3cm}

\section{Conclusion \& Limitation}
% \vspace{-0.2cm}
\quad In this paper, we propose UniHDA, a unified and versatile framework for multi-modal hybrid domain adaptation. To enable multiple modalities, we leverage CLIP encoder to project the references into a unified embedding space. For hybrid domain, we demonstrate the compositional capabilities of direction vectors in CLIP’s embedding space and linearly interpolate direction vectors of multiple target domains. In addition, we propose a new cross-domain spatial structure loss to improve consistency, which is conducted in generator-agnostic space and versatile for various generators. We believe our work is an important step towards generative domain adaptation, since we have demonstrated the source generator can be effectively adapted to a hybrid domain with multi-modal references and maintain robust cross-domain consistency. Our code will be made public.

% \noindent
% \textbf{Limitation.} 
While UniHDA effectively realizes multi-modal hybrid domain adaptation, it also has the limitation. To encode both image and text into a shared embedding space, we utilize pre-trained CLIP during training time, which might bring potential bias for some domains. Nevertheless, we believe that the exploration of the novel task is significant for future work and solutions could be integrated into UniHDA to eliminate the bias.
\label{conclusion}

% \clearpage  % TODO REVIEW/FINAL: This \clearpage needs to be removed from both review and camera-ready versions.

\bibliographystyle{splncs04}
\bibliography{main}

\end{document}

% \title{UniHDA: Towards Universal Hybrid Domain Adaptation of Image Generators} 
\title{UniHDA: A Unified and Versatile Framework for Multi-Modal Hybrid Domain Adaptation Supplementary Material}

\titlerunning{UniHDA}

% % TODO FINAL: Replace with your author list. 
% % Include the authors' OCRID for the camera-ready version, if at all possible.
% Second Author\inst{2,3}\orcidlink{1111-2222-3333-4444} \and
% Third Author\inst{3}\orcidlink{2222--3333-4444-5555}}

% % TODO FINAL: Replace with an abbreviated list of authors.
\authorrunning{Hengjia Li et al.}

\author{Hengjia Li$^{1}$ \and Yang Liu$^{1}$ \and Yuqi Lin$^{1}$\and Zhanwei Zhang$^{1}$\and Yibo Zhao$^{1}$\and Weihang Pan$^{1}$\and Tu Zheng$^{2}$\and Zheng Yang$^{2}$\and Chunjiang Yu$^{3}$\and Boxi Wu$^{1}$\and Deng Cai$^{1}$ 
}
\institute{
\textsuperscript{1} {Zhejiang University} \;
\textsuperscript{2} {Fabu Inc.} \;
\textsuperscript{3} {Ningbo Port} \;
}

\maketitle

% \begin{center}
% \centering
% \includegraphics[width=\linewidth]{fig/more_ablation.pdf} 
% \captionof{figure}{More qualitative results to verify the effectiveness of our proposed $\mathcal{L}_{\text{CSS}}$.
% }
% \label{fig:more_ablation}
% \end{center}

% \begin{figure*}[ht]
% \centering
% \includegraphics[width=.95\linewidth]{fig/3d.pdf}
% % \vspace{-0.2cm}
% \caption{Hybrid domain adaptation in 3D generator. To show the versatility of UniHDA, we apply it on the popular 3D-aware generator, EG3D~\cite{chan2022efficient}.}
% \label{fig:3d}
% % \vspace{-0.2cm}
% \end{figure*}

\begin{figure*}[ht]
\centering
\includegraphics[width=\linewidth]{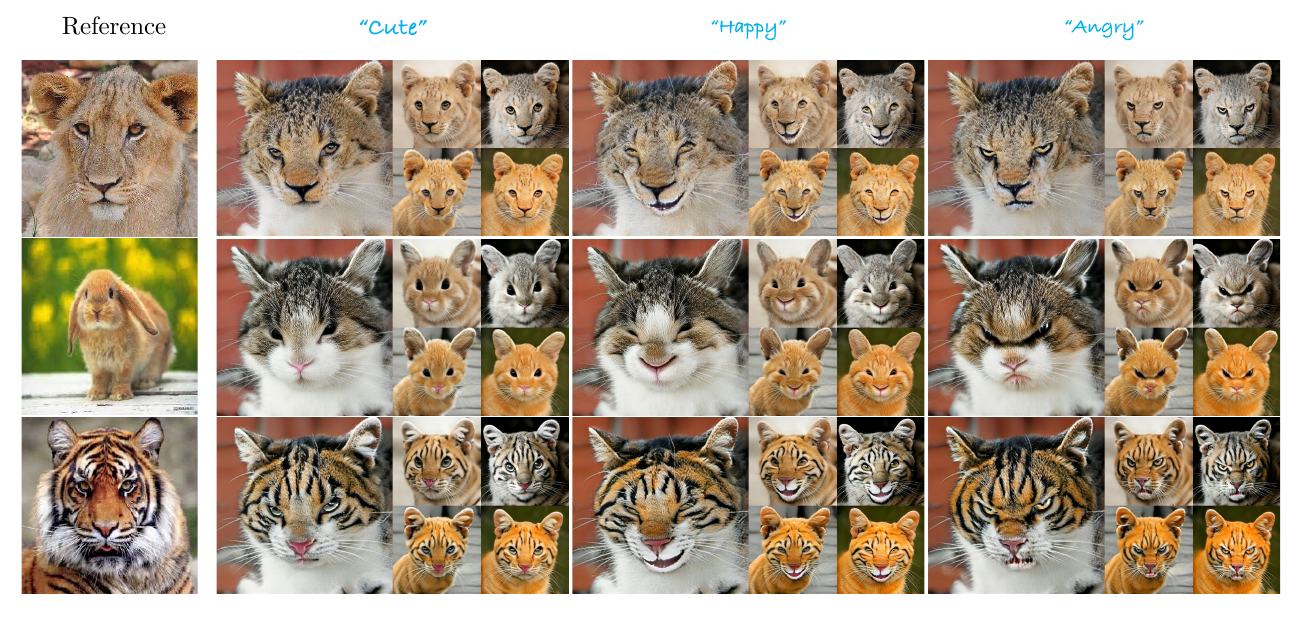}
% \vspace{-0.2cm}
\caption{Hybrid domain adaptation from AFHQ \textit{cat} to incompatible domains, \ie \textit{lion}, \textit{rabbit}, and \textit{tiger}.}
\label{fig:cross-cat}
% \vspace{-0.2cm}
\end{figure*}

\begin{figure*}[ht]
\centering
\includegraphics[width=\linewidth]{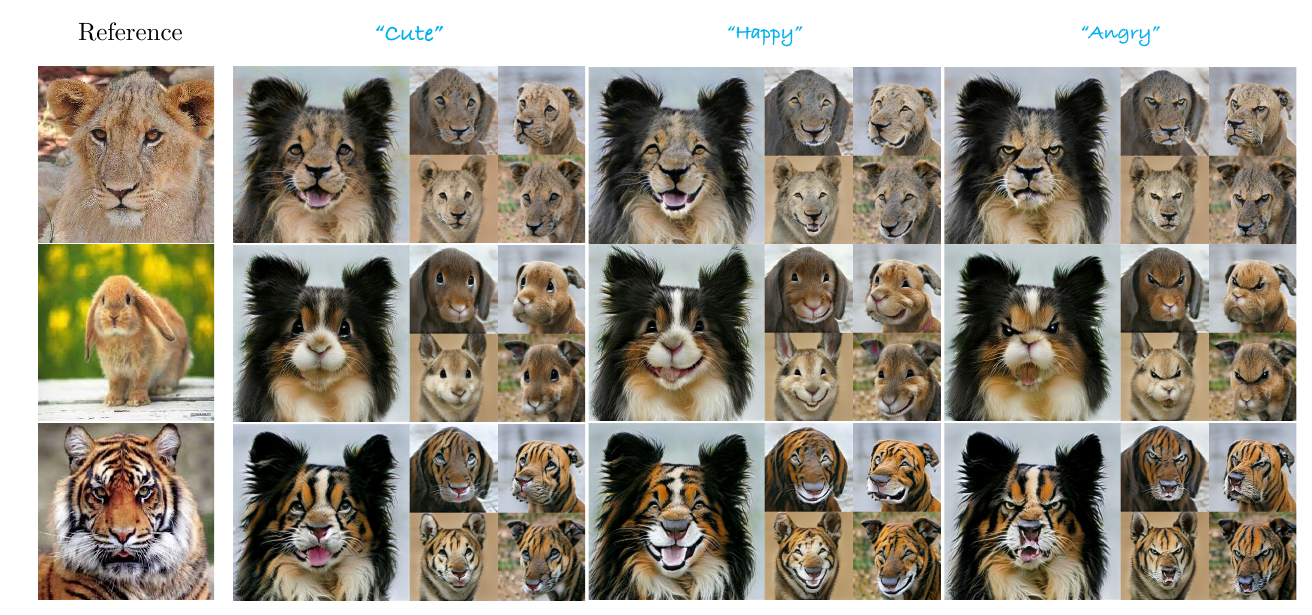}
% \vspace{-0.2cm}
\caption{Hybrid domain adaptation from AFHQ  \textit{dog} to incompatible domains, \ie \textit{lion}, \textit{rabbit}, and \textit{tiger}.}
\label{fig:cross-dog}
% \vspace{-0.2cm}
\end{figure*}

\begin{figure*}[ht]
\centering
\includegraphics[width=\linewidth]{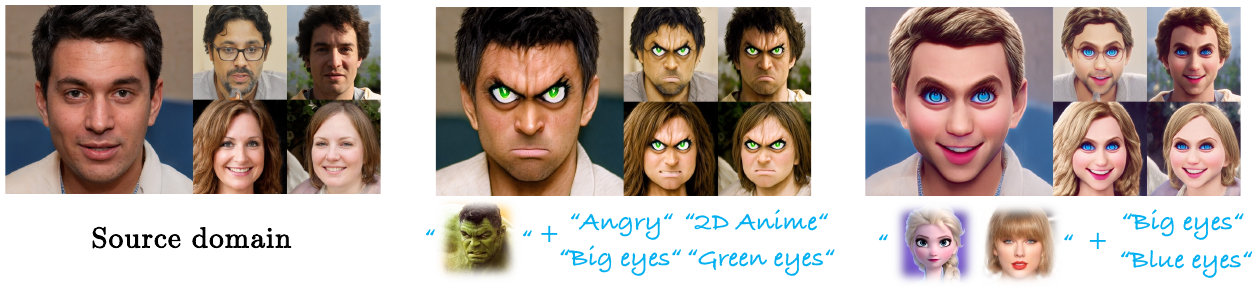}
% \vspace{-0.2cm}
\caption{The results of hybrid domain adaptation from FFHQ to the hybrid of more domains.}
\label{fig:more_domain}
% \vspace{-0.2cm}
\end{figure*}

\begin{figure*}[ht]
\centering
\includegraphics[width=\linewidth]{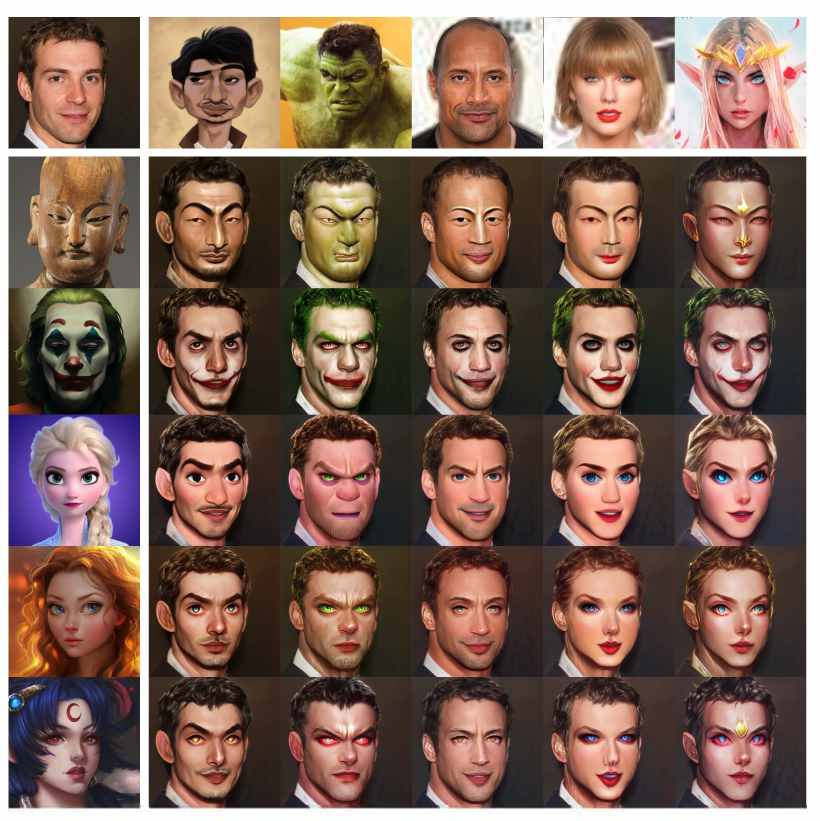}
% \vspace{-0.2cm}
\caption{More results of image-image hybrid domain adaptation. The source image is in the top-left corner, and the first row and column consist of training images.}
\label{fig:all-img-img}
% \vspace{-0.2cm}
\end{figure*}

\begin{figure*}[ht]
\centering
\includegraphics[width=\linewidth]{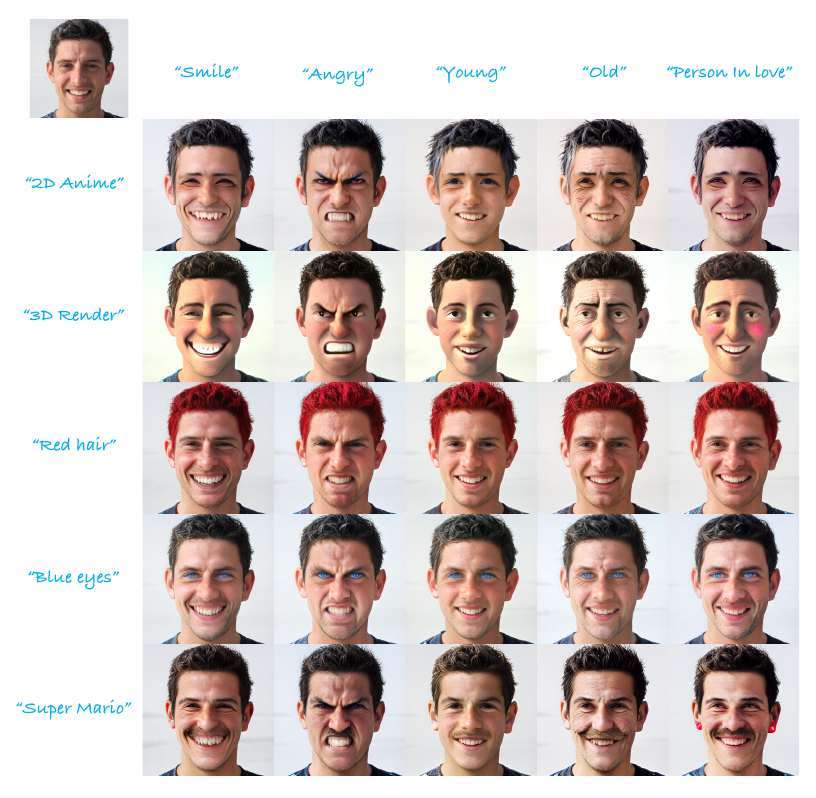}
% \vspace{-0.2cm}
\caption{More results of text-text hybrid domain adaptation. The source image is in the top-left corner, and the first row and column consist of text prompts.}
\label{fig:all-text-text}
% \vspace{-0.2cm}
\end{figure*}

\begin{figure*}[ht]
\centering
\includegraphics[width=\linewidth]{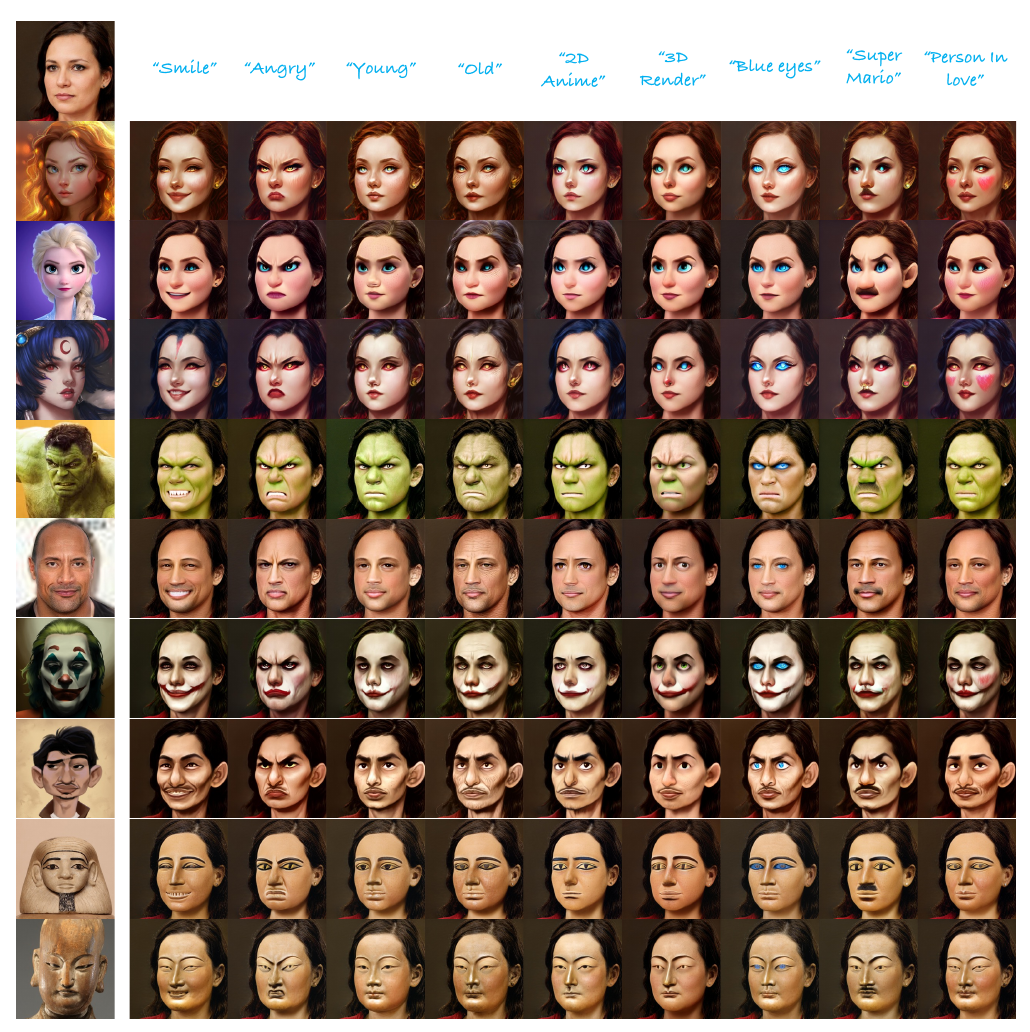}
% \vspace{-0.2cm}
\caption{More results of image-text hybrid domain adaptation. The source image is in the top-left corner. The first row and column consist of training images and text prompts respectively.}
\label{fig:all-img-text}
% \vspace{-0.2cm}
\end{figure*}

\begin{figure*}[ht]
\centering
\includegraphics[width=\linewidth]{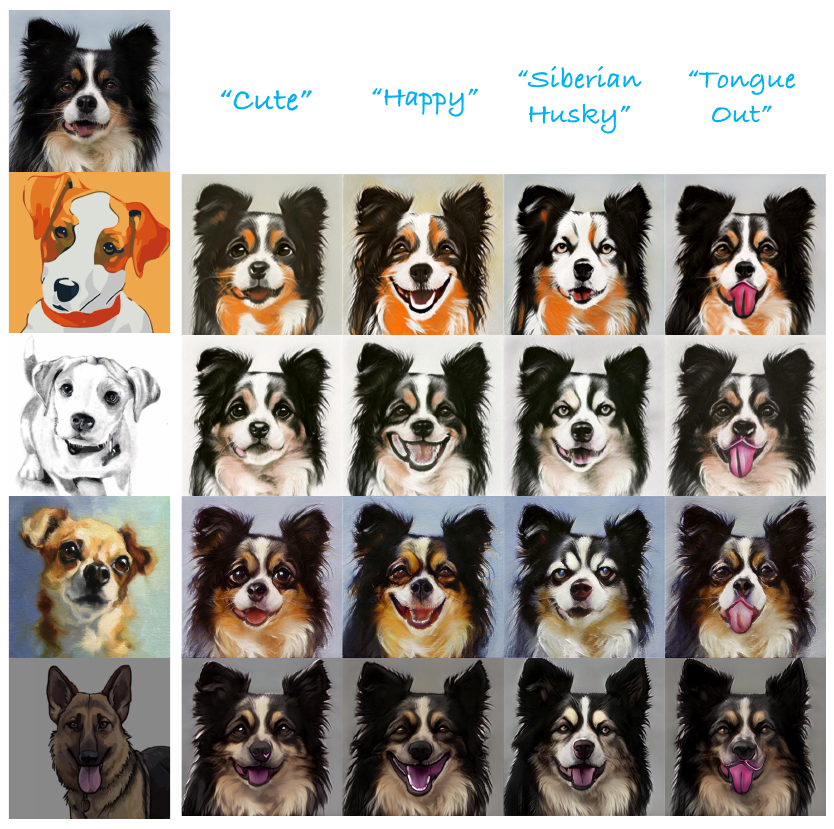}
% \vspace{-0.2cm}
\caption{More results of image-text hybrid domain adaptation on AFHQ \textit{dog}.}
\label{fig:more-dog}
% \vspace{-0.2cm}
\end{figure*}

\begin{figure*}[ht]
\centering
\includegraphics[width=\linewidth]{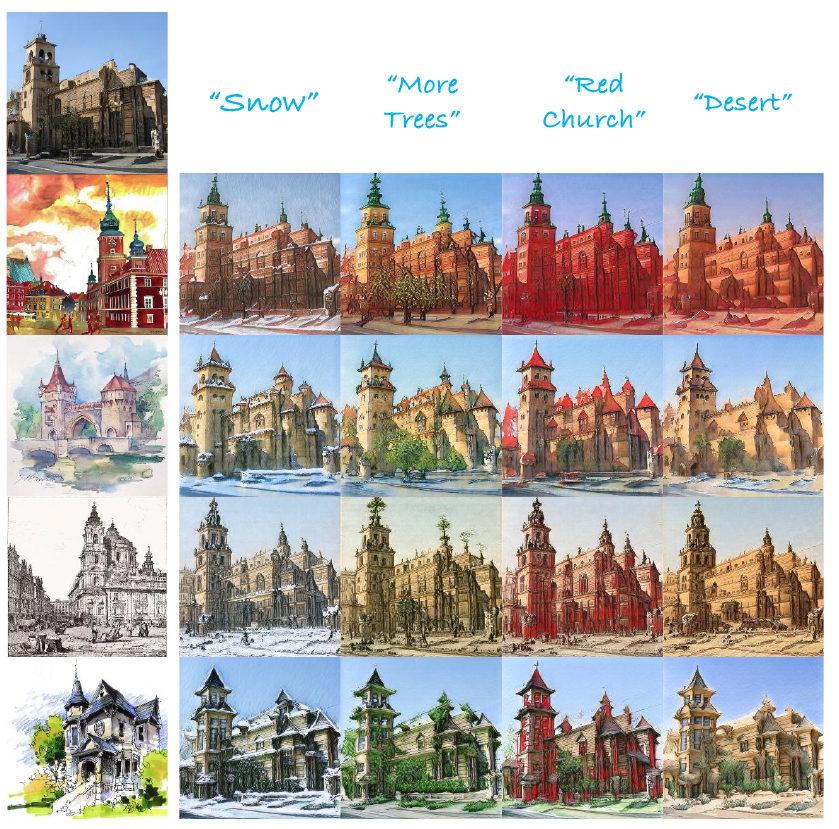}
% \vspace{-0.2cm}
\caption{More results of image-text hybrid domain adaptation on LSUN \textit{church}.}
\label{fig:more-church}
% \vspace{-0.2cm}
\end{figure*}

\begin{figure*}[ht]
\centering
\includegraphics[width=\linewidth]{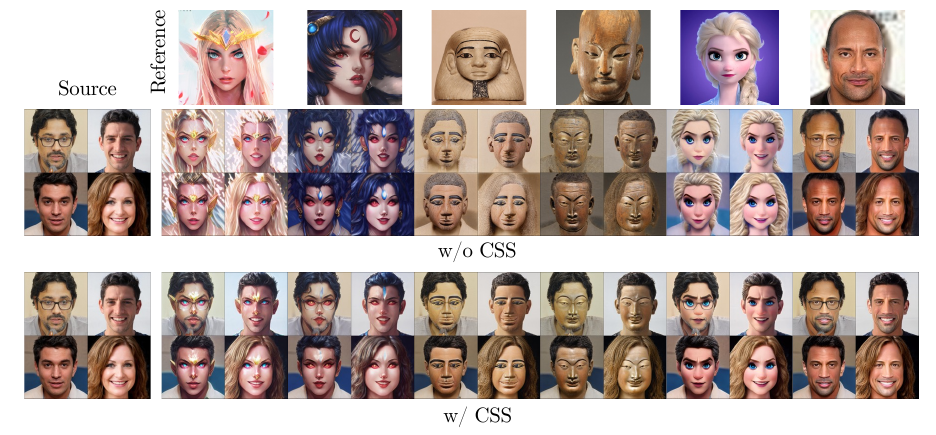} 
\captionof{figure}{More qualitative results to verify the effectiveness of our proposed $\mathcal{L}_{\text{CSS}}$.
}
\label{fig:more_ablation}
\end{figure*}

\begin{figure*}[t]
\centering
\includegraphics[width=\linewidth]{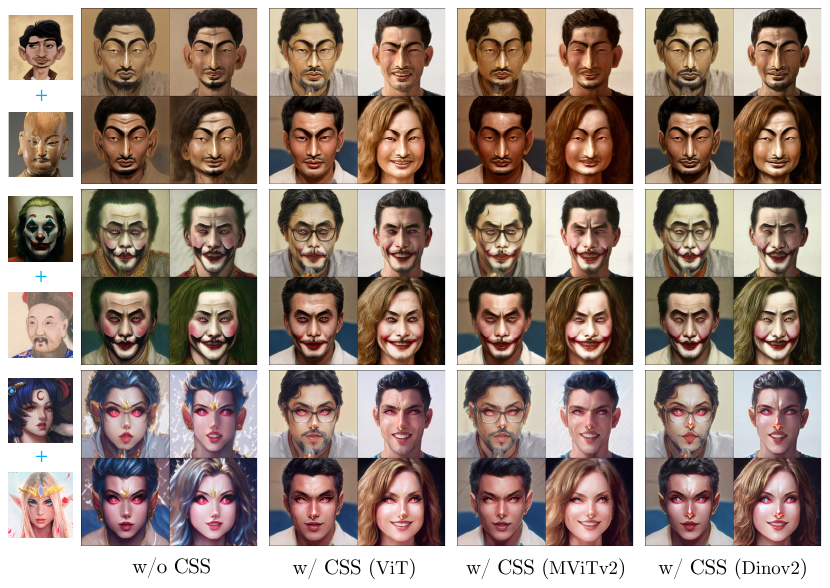}
% \vspace{-0.2cm}
\caption{Effect of different pre-trained image encoders for CSS on image-image hybrid domain adaptation.}
\label{fig:encoder-ii}
% \vspace{-0.2cm}
\end{figure*}

\begin{figure*}[t]
\centering
\includegraphics[width=\linewidth]{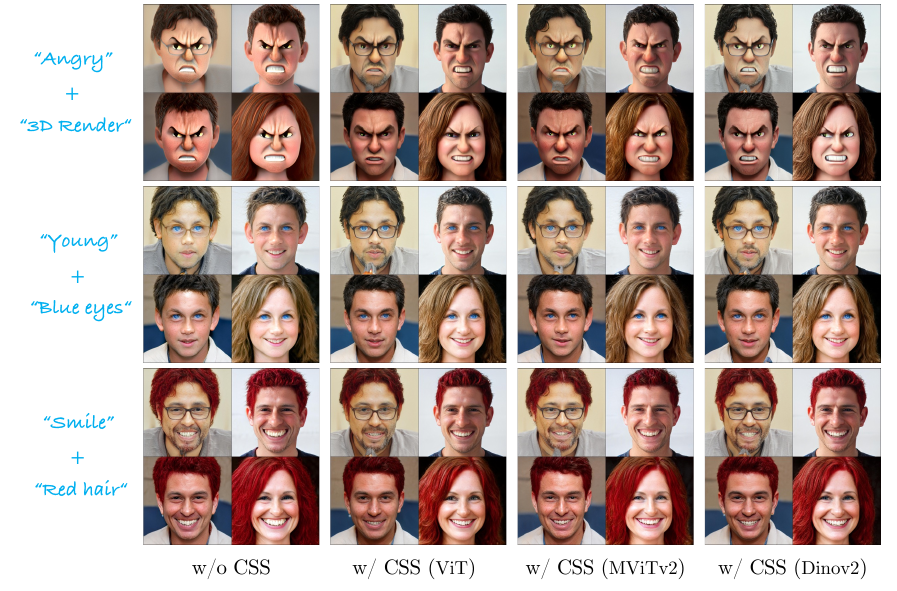}
\vspace{-0.2cm}
\caption{Effect of different pre-trained image encoders for CSS on text-text hybrid domain adaptation.}
\label{fig:encoder-tt}
\vspace{-0.2cm}
\end{figure*}

% \begin{figure*}[t]
% \centering
% \includegraphics[width=.9\linewidth]{fig/encoder-it.pdf}
% \vspace{-0.2cm}
% \caption{Effect of different pre-trained image encoders for CSS on image-text hybrid domain adaptation.}
% \label{fig:encoder-it}
% \vspace{-0.2cm}
% \end{figure*}

\begin{figure*}[ht]
\centering
\includegraphics[width=\linewidth]{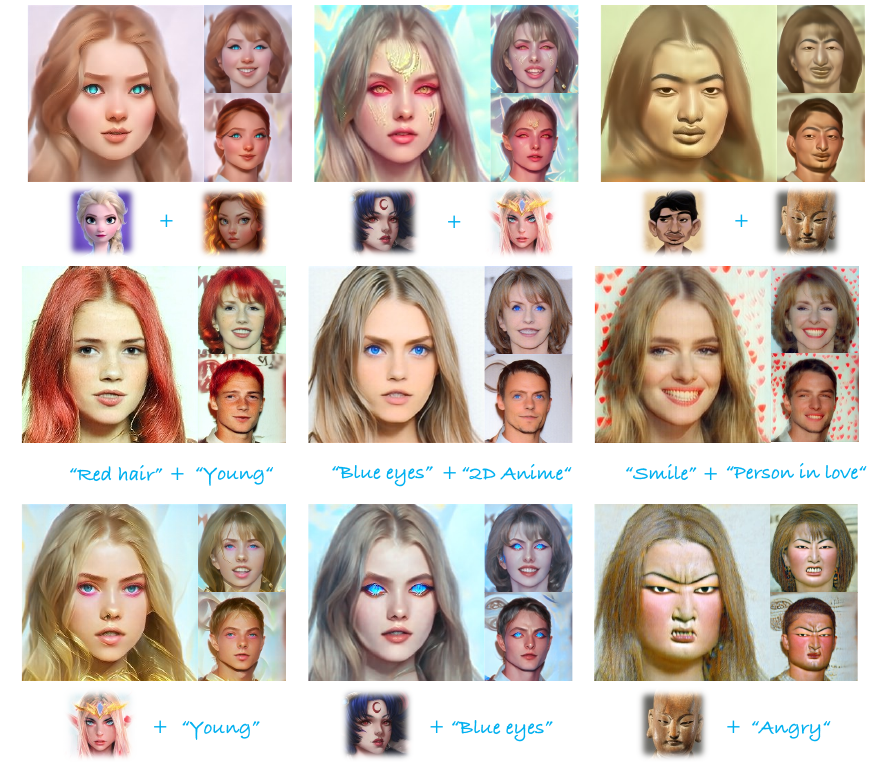}
% \vspace{-0.2cm}
\caption{More results of UniHDA with DiffusionCLIP.}
\label{fig:more-diff}
% \vspace{-0.2cm}
\end{figure*}

\section{Appendix}
% \subsection{Hybrid Domain Adaptation of 3D Generator}
% \quad To verify the versatility of UniHDA, We apply it on the popular 3D-aware image generation method, EG3D~\cite{chan2022efficient}. Specifically, we replace the discrimination loss with our proposed $\mathcal{L}_{direct}$ and $\mathcal{L}_{\text{CSS}}$ for hybrid domain adaptation. As shown in \cref{fig:3d}, the results effectively integrate the attributes and preserve the characters and poses of source domain.

\subsection{Results of Incompatible Domain Adaptation}
\quad To verify the generalization ability of UniHDA, we additionally conduct experiments for hybrid domain adaptation on incompatible domains, \ie from \textit{cat} to \textit{rabbit}. As shown in \cref{fig:cross-cat} and \cref{fig:cross-dog}, we start from AFHQ~\cite{choi2020stargan} \textit{cat} and \textit{dog}   to incompatible domains respectively. We can observe that the results effectively integrate the attributes of the corresponding domain and maintain robust consistency with source domain.

\subsection{More Qualitative Results}
\label{append:qual}
\quad We apply UniHDA to adapt the generator on FFHQ~\cite{karras2019style} to more hybrid domains, \ie, text-text, image-image, and image-text. As shown in \cref{fig:more_domain}, \cref{fig:all-img-img}, \cref{fig:all-text-text}, and \cref{fig:all-img-text}, UniHDA successfully generates images with integrated characteristics from multiple target domains and maintains robust consistency with the source domain. Besides, we showcase more results of hybrid domain adaptation from AFHQ \textit{dog} and LSUN \textit{church}~\cite{yu2015lsun} in \cref{fig:more-dog} and \cref{fig:more-church}.

\subsection{More Ablation of CSS Loss}
\label{append:ablation}
\quad As depicted in Sec. 4.8 of the main paper, our proposed $\mathcal{L}_{\text{CSS}}$ significantly alleviates overfitting and improves cross-domain consistency.  Results with $\mathcal{L}_{\text{CSS}}$ achieve better SCS score, indicating that they maintain stronger consistency with the source domain. Additionally, we show more qualitative results in \cref{fig:more_ablation} to verify the effectiveness of UniHDA.

\subsection{More Ablation of Encoder for CSS}
\quad In Sec. 4.9 of the main paper, we conduct experiments on pre-trained ViT, MViTv2, and Dinov2 to explore the impact of different image encoders for CSS. Additionally, we showcase the results of image-image and text-text in \cref{fig:encoder-ii} and \cref{fig:encoder-tt} respectively. We can observe that all of them improve the consistency with source domain compared with the baseline approach. Furthermore, they exhibit a similar qualitative style, which demonstrates that our CSS is agnostic to different pre-trained image encoders.

\subsection{More Results for DiffusionCLIP}
\quad To demonstrate the versatility of UniHDA, we apply it on DiffusionCLIP in Sec. 4.7 in the main paper. As shown in \cref{fig:more-diff}, we showcase more results for DiffusionCLIP including image-image, text-text, and image-text. All results achieve hybrid domain adaptation and preserve strong corss-domain consistency.

\subsection{Implement Details}
\label{append:detail}
\quad Following the setting of previous generative domain adaptation  methods~\cite{gal2021stylegan,nitzan2023domain}, we utilize the batch size of 4 and ADAM Optimizer with a learning rate of 0.002 for all experiments during training. A training session typically requires 300 iterations in 2 minutes, which significantly reduces training time compared with adversarial methods for generative domain adaptation. Note that we conduct all experiments on a single NVIDIA RTX 4090 GPU. \textcolor{blue}{The code will be open source.}

For experiments on FFHQ, we generate images with 1024 × 1024 resolution. As for AFHQ-Dog and LSUN-Church, we operate on 512 × 512 and 256 × 256 resolution images respectively.

% \subsection{Rationale}
% \label{sec:rationale}
% % 
% Having the supplementary compiled together with the main paper means that:
% % 
% \begin{itemize}
% \item The supplementary can back-reference sections of the main paper, for example, we can refer to \cref{sec:intro};
% \item The main paper can forward reference sub-sections within the supplementary explicitly (e.g. referring to a particular experiment); 
% \item When submitted to arXiv, the supplementary will already included at the end of the paper.
% \end{itemize}
% % 
% To split the supplementary pages from the main paper, you can use \href{https://support.apple.com/en-ca/guide/preview/prvw11793/mac#:~:text=Delete%20a%20page%20from%20a,or%20choose%20Edit%20%3E%20Delete).}{Preview (on macOS)}, \href{https://www.adobe.com/acrobat/how-to/delete-pages-from-pdf.html#:~:text=Choose%20%E2%80%9CTools%E2%80%9D%20%3E%20%E2%80%9COrganize,or%20pages%20from%20the%20file.}{Adobe Acrobat} (on all OSs), as well as \href{https://superuser.com/questions/517986/is-it-possible-to-delete-some-pages-of-a-pdf-document}{command line tools}.    

% \clearpage  % TODO REVIEW/FINAL: This \clearpage needs to be removed from both review and camera-ready versions.

\bibliographystyle{splncs04}
\bibliography{main}